# IGADA-IoT: IoT Sensor Energy Optimization in Wireless Sensor Networks Driven by Automatic Data Augmentation

Mingchun Sun, Rongqiang Zhao*, Muhammad Abdul Munnaf, Jie Liu *Fellow, IEEE*

***Abstract*—In wireless sensor networks (WSNs), data augmentation is a novel method to improve sampling-frequency decision performance, thereby enabling energy optimization for IoT (Internet of Things) sensors. However, existing methods rely on a single generator and empirically determined quantities, failing to establish a mapping between dynamic information gaps and multiple generators, and overlooking the heterogeneity of generated samples. Moreover, an evaluation and a closed-loop method that jointly consider the information gap and the model performance are lacking. To address these issues, we propose an information gap-guided IoT sensor automatic data augmentation framework (IGADA-IoT) with hierarchical multi-generator collaboration and scheduling over multiple rounds. Capabilities of different generators are jointly utilized to reduce the information gaps. In the IGADA-IoT, a hierarchical multi-generator collaboration and scheduling strategy (HMGCS) is proposed to enhance the targetedness and rationality of generated sample allocation. An information gap–model performance joint evaluation and closed-loop method (IGMP-EC) is proposed to enhance the accuracy of augmentation decisions, and to mitigate the risks of under-augmentation and over-augmentation. Experimental results show that the IGADA-IoT improves the average accuracy of multiple downstream models by 7.28%. Compared with advanced data augmentation methods, the average accuracy is improved by 8.90%. Compared with the individual generators, the average accuracy is improved by 7.63%. Furthermore, public IoT sensor datasets from the UCR Archive and real-world deployments demonstrate the accuracy and deployment feasibility of the proposed method.**

## I. Introduction

In WSNs, IoT sensors serve as the foundation for sensing environmental information. Their energy consumption directly affects the sustainability of the sensor node and the reliability of its services [1], [2]. Due to dynamic operating conditions and limited energy resources, fixed-frequency sampling causes energy waste during stationary stages and information loss during non-stationary stages [3]. Sensor lifetime and acquired data information content are constrained without effective energy optimization [4]. Therefore, IoT sensor energy optimization is necessary.

*Corresponding author: Rongqiang Zhao@zhaorq@hit.edu.cn.

Mingchun Sun, Rongqiang Zhao, and Jie Liu are with the Faculty of Computing, Harbin Institute of Technology, Harbin 150001, China.

Mingchun Sun, Rongqiang Zhao, and Jie Liu are with the State Key Laboratory of Smart Farm Technologies and Systems, Harbin 150001, China.

Muhammad Abdul Munnaf is with the Agricultural Biosystems Engineering, Wageningen University and Research, Wageningen 6703 CH, Netherlands.

Existing methods for energy optimization mainly focus on adaptive sampling and energy-aware sensing [5]. They dynamically adjust the sensing behavior of IoT sensors according to data variations, device energy, or application requirements, thereby extending node lifetime [6]. The energy efficiency of the WSNs is also improved. Artificial intelligence algorithms are used to predict environmental changes. The changes are mapped to class labels, which are used to adjust the sensing intervals [7]. Energy consumption is minimized without compromising the integrity of the acquired data. Other methods adopt adaptive sampling intervals and policies to balance the energy constraints of IoT sensors with sensing objectives [8]. Moreover, combining edge intelligence with energy models ensures the long-term stable operation of IoT sensors and mitigates the risk of battery depletion [9], [10]. However, existing methods mostly emphasize the optimization of sensing and sampling, while paying insufficient attention to the data foundation. Class-imbalanced and scale-limited datasets still constrain the sampling-frequency decision. Therefore, data augmentation is essential for further enabling the performance of IoT sensor energy optimization.

IoT sensor signals inherently take the form of time series. Data augmentation for time series generally includes data-level augmentation and model-level augmentation [11]. Data-level augmentation employs predefined strategies to perform mathematical transformations on the acquired data, such as interpolation, decomposition, recombination, and perturbation [12], [13]. Data diversity is increased and local representations are improved. Model-level augmentation employs generative models to learn the underlying data distribution and then generate statistically plausible data, such as generative adversarial networks (GANs), variational autoencoders (VAEs), and diffusion models [14], [15]. Scarce representations and missing information in the data are supplemented [16], [17]. However, data augmentation for energy optimization relies on a single generator and empirically determined quantities. Capabilities of different generators and the information gaps in datasets are overlooked, failing to establish a mapping between dynamic information gaps and multiple generators. Generated data are treated as homogeneous resources, overlooking the heterogeneity of generated samples. Furthermore, an evaluation and a closed-loop method that jointly consider the information gap and the model performance are lacking.

The motivation of this work is to address these limitations. Class-imbalanced and scale-limited acquired datasets can be augmented to reduce the information gaps automatically. The accuracy and generalizability of the sampling-frequency decision models are improved, which further enables the energy optimization of IoT sensors in real-world deployments. To address these issues, we propose an information gap-guided IoT sensor automatic data augmentation framework (IGADA-IoT) with hierarchical multi-generator collaboration and scheduling over multiple rounds. The proposed method includes a hierarchical multi-generator collaboration and scheduling strategy (HMGCS) and an information gap–model performance joint evaluation and closed-loop method (IGMP-EC). Through the IGMP-EC, the HMGCS performs targeted and quantitative budget allocation during the collaboration and scheduling of multiple generators in each round, thereby forming an automatic data augmentation closed loop. Unlike existing data augmentation methods that rely on a single generator and empirically determined quantities, the proposed method jointly utilizes the capabilities of generators and the information gaps in the dataset. Unlike homogeneous data allocation strategies, the proposed method allocates generated data to each class precisely according to the round-wise information gaps and the downstream model performance. Unlike open-loop augmentation methods that lack benefit evaluation, the proposed method enables the augmentation to automatically terminate when the marginal gain in the performance is exhausted. The contributions are summarized as follows:

1) We propose the IGADA-IoT for automatic data augmentation in WSNs. Capabilities of different generators are jointly utilized to reduce the information gaps.

2) We propose the HMGCS to quantify the information gap, which is used to collaborate and schedule multiple generators in class-level, generator-level, and budget-level for augmentation. The targetedness and rationality of generated sample allocation are enhanced. Differentiated scheduling demands are satisfied.

3) We propose the IGMP-EC to evaluate and perform accept-or-reject decisions on candidate augmentation over multiple rounds. The accuracy of augmentation decisions is enhanced, and the risks of under-augmentation and over-augmentation are mitigated.

The organization of the paper is as follows. In Section II, the related works are presented and discussed. In Section III, the methodology and theory are elaborated in detail. In Section IV, the experimental results are shown and analyzed, and conclusions are drawn in Section V.

## II. Related Works

### A. Energy Optimization for IoT Sensors

Existing energy optimization methods for IoT sensors can be divided into two categories. The first category focuses on the sensing task itself and optimizes energy by adjusting the sensing interval or sampling frequency [18], [19]. Classical energy-aware adaptive sampling adjusts sampling frequencies in response to signal variations or battery conditions to balance sensing quality and energy consumption [20], [21]. Recently, machine learning models are used to learn the variation patterns of sensing data and dynamically adjust the sensing interval, such as logistic regression (LR), random forest (RF), and support vector machine (SVM) [22], [23]. In WSNs, the effectiveness of machine learning models in energy optimization and node lifetime extension has been demonstrated [24], [25]. Furthermore, machine learning models are also used to regulate adaptive sampling frequency for establishing a better balance between sensing quality and energy consumption. As the environmental complexity increases, the coupling among the multi-dimensional features of data becomes more pronounced. Machine learning models are gradually limited in representing high-dimensional dynamic patterns, deep learning models therefore receive increasing attention [26]. Deep learning models are widely used for energy optimization, such as convolutional neural network (CNN), long short-term memory (LSTM), gated recurrent unit (GRU), and Transformer [27], [28], [29]. Temporal dependencies, nonlinear relationships, and cross-dimensional feature interactions are explored more deeply, thereby providing stronger modeling capability for sampling frequency decision and energy optimization. The second category places more emphasis on system-level energy optimization, such as communication transmission, computation offloading, and clustering-based routing [30], [31]. The methods reduce system energy consumption by optimizing data transmission, task allocation, and network organization, which rely on adjustments to system control logic, network protocols, or even infrastructure deployment [32], [33]. In contrast, the first category enables energy optimization at the data source in a lighter and less intrusive manner, without changing the existing architecture. Although it offers greater deployment flexibility and practical advantages in resource-constrained WSNs, the acquired datasets for IoT sensors commonly suffer from class imbalance and limited scale.

### B. Data Augmentation

For time series, the core of data-level augmentation lies in predefined strategies, which expand the dataset and preserve its essential semantic features. Recursive interpolation method (RIM) generates data through recursive interpolation [34], and nonsignificant subsequence interval interpolation method (Non-SIM) further incorporates physical priors to enable directed generation [11]. Dynamic time warping (DTW) performs guided temporal warping by using samples from the same class as references [35], time-series warping (TSW) simulates local pattern deformation by alternately compressing and stretching subsequences [36]. Synthetic minority oversampling technique (SMOTE) and its variant, Borderline-SMOTE, are also interpolation-based augmentation methods [37], [38]. Automated augmentation methods learn suitable policies or transformations for data augmentation through differentiable optimization [39], [40]. Although data-level augmentation methods are convenient to implement and lightweight, they are constrained by predefined strategies and are limited in characterizing the coupling relationships among multiple variables. Therefore, model-level augmentation methods gradually become dominant [41]. The core of model-level augmentation lies in generative models, including generative adversarial networks (GANs), variational autoencoders (VAEs), and diffusion models [42], [43]. However, GANs suffer from unstable training and are prone to mode collapse. Data generated by VAEs are often overly smooth and lack sufficient detail. Due to their stable training, superior characterization of complex data distributions, and high sample diversity, diffusion models have emerged as the leading generative model [44]. To make the generation controllable, DS-Diffusion, Diffusion-TS, and MG-TSD incorporate sequence patterns or physical priors as conditions to guide diffusion model training [44], [45], [46]. Furthermore, ImagenTime bridges time-series diffusion and image diffusion, enabling the application of advanced image-generative models to sequential data [47], [48]. Both generation strategies and generative models are widely used in data augmentation. Multi-generator architectures also support time-series generation, but their generators mainly perform distinct component-generation tasks tailored to specific application domains [49], [50]. Generator allocation in multi-generator architectures is not explicitly linked to evolving information gaps. Although existing methods for data augmentation are already capable of generating high-fidelity data, many rely on a single generator and empirically determined quantities. Such methods fail to establish a mapping between dynamic information gaps and multiple generators, and overlook the heterogeneity of generated samples. Moreover, an evaluation and a closed-loop method that jointly consider the information gap and the model performance are lacking.

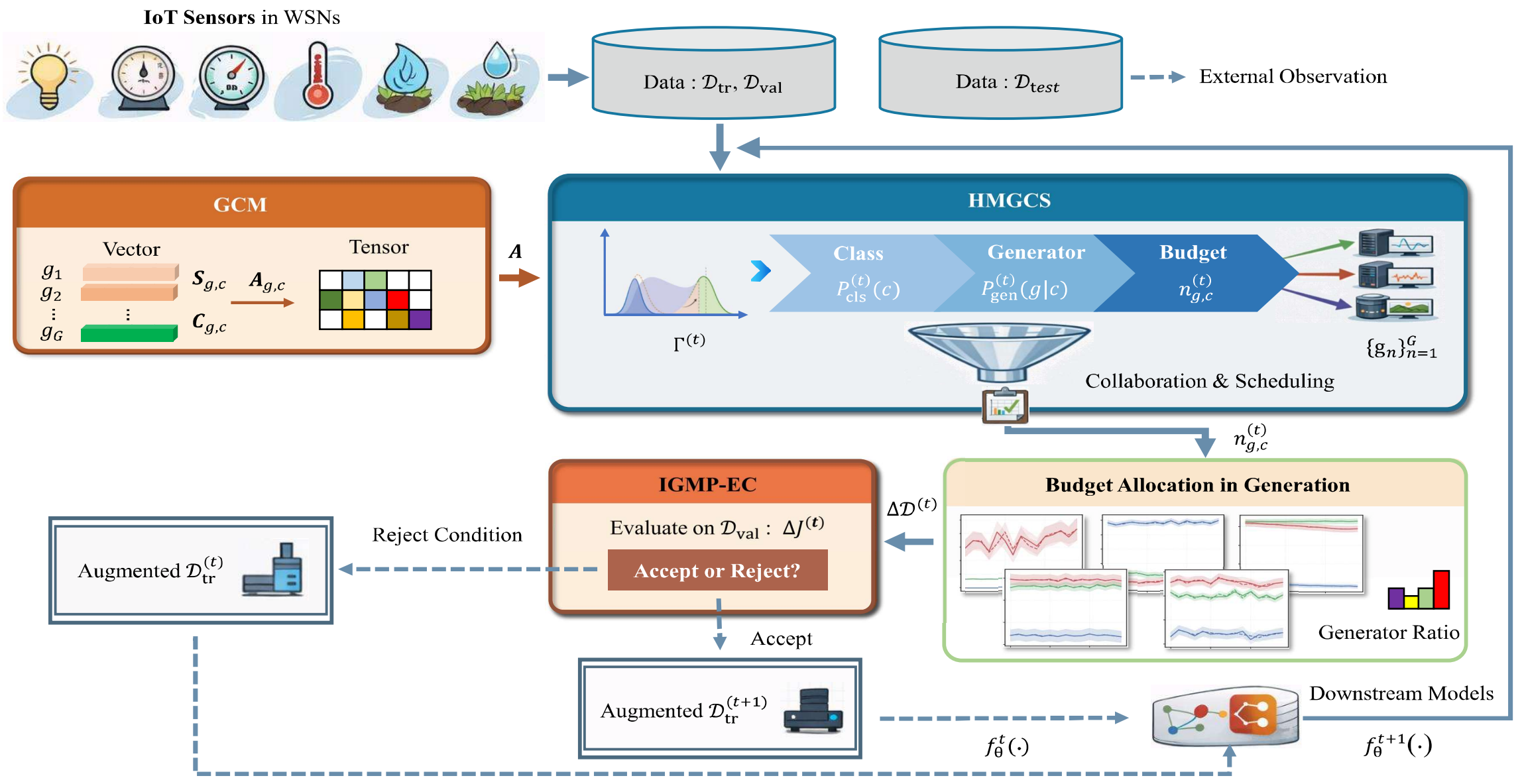

Fig. 1. Overview of IGADA-IoT.

## III. Methodology

We consider an automatic data augmentation framework in WSNs for energy optimization. The data acquired by IoT sensors are represented as multivariate time series. The data space is defined as $\mathcal{X} \subseteq \mathbb{R}^{T\times F}$, where $F$ denotes the number of features, $T$ denotes the time window length. The energy optimization task is formulated as a sampling-frequency decision task. Data augmentation is used to improve the performance of sampling-frequency classification models in the sampling-frequency decision task, where the predicted labels are mapped to predefined sampling intervals. These models are used to optimize sampling intervals and the number of sampling events, thereby affecting sampling energy consumption. The downstream models map the acquired data to one of the discrete sampling-frequency decision labels in the label space $\mathcal{Y}$. The IGADA-IoT is proposed to collaborate and schedule multiple generators in multi-round automatic data augmentation. The class-imbalanced and scale-limited IoT sensor dataset is augmented to improve the downstream model performance in sampling-frequency decision. The IGADA-IoT incorporates generator capability modeling (GCM), the HMGCS, and the IGMP-EC. An overview of the IGADA-IoT is shown in Fig. 1. The following subsections will introduce the GCM, the HMGCS, and the IGMP-EC in detail.

### A. GCM

The GCM is proposed to tensorize the capabilities of multiple generators for different classes into a capability tensor. The acquired data are split into a training set $\mathcal{D}_{\rm tr}$, a validation set $\mathcal{D}_{\rm val}$, and a test set $\mathcal{D}_{\rm test}$, where $\boldsymbol{x}_i$ denotes a multivariate time window of length $T$, and $c \in \mathcal{Y}$ denotes the class. $C$ is the number of classes. To prevent information leakage, only $\mathcal{D}_{\rm tr}$ is used for data augmentation. $g \in \{g_n\}_{n=1}^{G}$ denotes the generators. $G$ is the number of generators. Assume that the real data in $c$ and the data generated by generator $g$ are denoted as $\mathcal{X}_r^{(c)} = \{\boldsymbol{x}_{r,i}^{(c)}\}_{i=1}^{N_r^{(c)}}$ and $\mathcal{X}_g^{(c)} = \{\boldsymbol{x}_{g,j}^{(c)}\}_{j=1}^{N_g^{(c)}}$ in $\mathcal{D}_{\rm tr}$. $N_r^{(c)}$ and $N_g^{(c)}$ denote their sample sizes, $i$ and $j$ are the indices. First, $\mathcal{X}_r^{(c)}$ and $\mathcal{X}_g^{(c)}$ are mapped into a class-conditional low-dimensional representation space. Consider that $\mathcal{X}$ represents two-dimensional temporal data, its vectorized representation is $\bar{\boldsymbol{x}} \in \mathbb{R}^{TF}$. $\bar{\boldsymbol{x}}$ is used to transform both real data and generated data into low-dimensional representations $\boldsymbol{z}_{r,i}^{(c)}$ and $\boldsymbol{z}_{g,j}^{(c)}$ under the class condition, as shown in Eq. (1).

$$\boldsymbol{z}_{r,i}^{(c)} = \phi_c(\bar{\boldsymbol{x}}_{r,i}^{(c)}), \qquad \boldsymbol{z}_{g,j}^{(c)} = \phi_c(\bar{\boldsymbol{x}}_{g,j}^{(c)}) \tag{1}$$

where $\phi_c(\cdot)$ denotes an embedding mapping composed of PCA-based dimensionality reduction and normalization. The low-dimensional representations reduce the redundancy of high-dimensional temporal data. They also avoid the interference of excessive original dimensionality and inconsistent scales in mutual information estimation.

Second, joint samples are constructed through a unified pairing operator $\Pi_{g,c}$. Different data augmentation methods correspond to different ways of constructing joint data. For augmentation strategies, real and generated data preserve explicit sample-level correspondence. For generative models, generated data are treated as a set independently sampled within a given class. The different methods reflect the sample provenance of the two method types without artificially imposing or altering sample correspondences. Therefore, $\Pi_{g,c}$ places different generators under the same distribution, as shown in Eq. (2).

$$\Pi_{g,c} : \left(\mathcal{Z}_r^{(c)}, \mathcal{Z}_g^{(c)}\right) \mapsto \mathcal{U}_{g,c} \tag{2}$$

where $\mathcal{U}_{g,c}$ denotes the joint set used for mutual information estimation. $\mathcal{Z}_r^{(c)} = \{\boldsymbol{z}_{r,i}^{(c)}\}_{i=1}^{N_r^{(c)}}$ and $\mathcal{Z}_g^{(c)} = \{\boldsymbol{z}_{g,j}^{(c)}\}_{j=1}^{N_g^{(c)}}$ denote the low-dimensional representation sets of real data and generated data in $c$, respectively. For augmentation strategies, $\Pi_{g,c}$ adopts sequential alignment pairing. For generative models, $\Pi_{g,c}$ adopts independent random pairing within the same class.

Third, generator capability tensor $\boldsymbol{A}$ is tensorized. Class-conditional mutual information is used to measure the information overlap between $\mathcal{X}_r^{(c)}$ and $\mathcal{X}_g^{(c)}$, as shown in Eq. (3).

$$I_{g,c} = I(\boldsymbol{Z}_r; \boldsymbol{Z}_g \mid \mathcal{Y} = c). \tag{3}$$

where $\boldsymbol{Z}_r$ and $\boldsymbol{Z}_g$ are the random variables constructed from $\mathcal{Z}_r^{(c)}$ and $\mathcal{Z}_g^{(c)}$ via $\Pi_{g,c}$. Kraskov–Stögbauer–Grassberger (KSG) nearest-neighbor estimator is used to estimate the mutual information, as shown in Eq. (4).

$$\widehat{I}_{g,c} = \psi(k) + \psi(n) - \frac{1}{n}\sum_{i=1}^{n}\Big(\psi(n_{\boldsymbol{x},i}+1) + \psi(n_{y,i}+1)\Big) \tag{4}$$

where $\psi(\cdot)$ denotes the digamma function, $n_{\boldsymbol{x},i}$ and $n_{y,i}$ denote the neighbor statistics of the $i$-th joint sample in the two marginal spaces, respectively. Since $\widehat{I}_{g,c}$ is affected by finite-sample bias, a permutation-based baseline correction is further used. Assume that $\widehat{I}_{g,c}^{raw}$ denotes the mutual information estimate under the original pairing, and $\widehat{I}_{g,c}^{perm}$ denotes the baseline mutual information obtained by randomly permuting the order of $\mathcal{X}_g^{(c)}$. The corrected mutual information is then defined as in Eq. (5).

$$\widehat{I}_{g,c}^{corr} = \max\Big(0,\ \mathbb{E}\big[\widehat{I}_{g,c}^{raw}\big] - \mathbb{E}\big[\widehat{I}_{g,c}^{perm}\big]\Big) \tag{5}$$

where $\mathbb{E}[\cdot]$ denotes the expectation operator, which is used to mitigate the randomness. To eliminate the influence of intra-class complexity differences on capability estimation, the class-conditional entropy of $\boldsymbol{Z}_r^{(c)}$ is computed. The similarity score $\boldsymbol{S}_{g,c}$ of $g$ for $c$ is defined in Eq. (6).

$$\boldsymbol{S}_{g,c} = \frac{\widehat{I}_{g,c}^{corr}}{H(\boldsymbol{Z}_r \mid \mathcal{Y} = c) + \varepsilon} \tag{6}$$

where Kozachenko–Leonenko (KL) nearest-neighbor differential entropy estimator is used and $\varepsilon$ is a stability term to prevent division by 0. $\boldsymbol{S}_{g,c}$ measures the proportion of class-conditional information overlap between $\mathcal{X}_r^{(c)}$ and $\mathcal{X}_g^{(c)}$. The contribution score $\boldsymbol{C}_{g,c}$ refers to the complementary or exploratory potential relative to the real distribution, as shown in Eq. (7).

$$\boldsymbol{C}_{g,c} = \boldsymbol{1} - \boldsymbol{S}_{g,c} \tag{7}$$

$\boldsymbol{S}_{g,c}$ and $\boldsymbol{C}_{g,c}$ are pairing-based heuristic scores and are treated as capability scores for collaborating and scheduling multiple generators. Based on $\boldsymbol{S}_{g,c}$ and $\boldsymbol{C}_{g,c}$, the capability of each generator is represented by a two-dimensional capability vector with a single degree of freedom, as shown in Eq. (8).

$$\boldsymbol{A}_{g,c} = [\boldsymbol{S}_{g,c},\ \boldsymbol{C}_{g,c}] \tag{8}$$

$\boldsymbol{S}_{g,c}$ and $\boldsymbol{C}_{g,c}$ are complementary representations of generator capability. A larger $\boldsymbol{S}_{g,c}$ indicates that $g$ is more suitable for providing refinement-oriented data that are close to the real distribution for $c$. A larger $\boldsymbol{C}_{g,c}$ indicates that $g$ is more suitable for providing exploratory data that expand the coverage space for $c$. The generator capability vectors form $\boldsymbol{A} \in \mathbb{R}^{G\times C\times 2}$, as shown in Eq. (9).

$$\boldsymbol{A} \leftarrow \{\boldsymbol{A}_{g,c}\}_{g=1,c=1}^{G,\ C} \tag{9}$$

It is worth noting that the construction of $\boldsymbol{A}_{g,c}$ presupposes a set of candidate generators that are trained or defined in advance. $\boldsymbol{A}_{g,c}$ serves as a scheduling prior within the IGADA-IoT, rather than as a standalone measure of the generated data quality. $\boldsymbol{A}$ tensorizes the capabilities of generators for different classes and is directly used in the HMGCS as the prior input. The algorithm flow of the GCM is described in Algorithm 1.

**Algorithm 1** GCM

**Require:** $\mathcal{D}_{\text{tr}}, \{g_n\}_{n=1}^{G}, \mathcal{Y}$
**Ensure:** $\boldsymbol{A}$

```
A ← 0 ∈ R^{G×C×2}
for c ∈ Y do
    for g ∈ {g_n}_{n=1}^G do
        X_r^(c), X_g^(c) ← D_tr, g, c
        Z_r^(c), Z_g^(c) ← φ_c(X_r^(c)), φ_c(X_g^(c)) using Eq. (1)
        U_{g,c} ← Π_{g,c}(Z_r^(c), Z_g^(c)) using Eq. (2)
        Î_{g,c}^raw ← U_{g,c} using Eqs. (3) and (4)
        Î_{g,c}^corr ← Î_{g,c}^raw using Eq. (5)
        S_{g,c}, C_{g,c} ← Î_{g,c}^corr using Eqs. (6) and (7)
        A_{g,c} ← S_{g,c}, C_{g,c} using Eq. (8)
    end for
end for
A ∈ R^{G×C×2} ← A_{g,c} using Eq. (9)
return A
```

## B. HMGCS

The HMGCS is proposed to quantify the information gap $\Gamma$ for collaborating and scheduling multiple generators in class-level, generator-level, and budget-level. First, the downstream model $f_\theta(\cdot)$ is trained on $\mathcal{D}_{\text{tr}}$ and validated on $\mathcal{D}_{\text{val}}$, and $\Gamma$ of $f_\theta(\cdot)$ on $\mathcal{D}_{\text{val}}$ is then quantified. To align data augmentation with the objective of the downstream task, $f_\theta(\cdot)$ and $\Gamma$ serve as feedback over multiple rounds in the HMGCS. $\mathcal{D}_{\text{test}}$, as an independent held-out sample from the target data distribution, is able to characterize the residual uncertainty or distributional mismatch without being affected by the model fitting process. Therefore, $\mathcal{D}_{\text{test}}$ is used for external observation only and does not participate in the multi-round automatic data augmentation. The automatic data augmentation is consistently restricted to $\mathcal{D}_{\text{tr}}$ and $\mathcal{D}_{\text{val}}$, and the information in $\mathcal{D}_{\text{test}}$ does not leak into the augmentation. Assume that $f_\theta^{(t)}(\cdot)$ denotes the downstream model in round $t$. For a sample $\boldsymbol{x}_i$ in $\mathcal{D}_{\text{val}}$, $f_\theta^{(t)}(\cdot)$ outputs a class probability vector, as shown in Eq. (10).

$$\boldsymbol{p}_i = f_\theta^{(t)}(\boldsymbol{x}_i) \tag{10}$$

In round $t$, a 4-dimensional information gap vector is defined. It consists of a sample size gap $h_{c,\text{size}}^{(t)}$, a distribution gap $h_{c,\text{dist}}^{(t)}$, a classification boundary gap $h_{c,\text{bdry}}^{(t)}$, and a prediction uncertainty gap $h_{c,\text{unc}}^{(t)}$, as shown in Eqs. (11)–(15).

$$\boldsymbol{h}_c^{(t)} = \left[h_{c,\text{size}}^{(t)}, h_{c,\text{dist}}^{(t)}, h_{c,\text{bdry}}^{(t)}, h_{c,\text{unc}}^{(t)}\right] \tag{11}$$

$$h_{c,\text{size}}^{(t)} = \max\left(0, \frac{N_{\max}^{ref} - N_c^{(t)}}{N_{\max}^{ref}}\right) \tag{12}$$

$$h_{c,\text{dist}}^{(t)} = 1 - \frac{1}{d_c}\sum_{q=1}^{d_c} \text{JS-sim}\big(p_{c,q}^{ref}, p_{c,q}^{(t)}\big) \tag{13}$$

$$h_{c,\text{bdry}}^{(t)} = 1 - \text{Rec}_c^{(t)} \tag{14}$$

$$h_{c,\text{unc}}^{(t)} = \frac{1}{|\mathcal{V}_c|}\sum_{i\in\mathcal{V}_c} \frac{H(\boldsymbol{p}_i)}{\log C} \tag{15}$$

$\mathcal{D}_{tr}$ is fixed. The data that are used to train $f_\theta^{(t)}(\cdot)$ in round $t$ form the current training set $\mathcal{D}_{tr}^{(t)}$. With increasing $t$, $\mathcal{D}_{tr}^{(t)}$ gradually includes more generated data. $N_c^{(t)}$ denotes the sample size of $c$ in $\mathcal{D}_{tr}^{(t)}$, and $N_{\max}^{ref}$ denotes the sample size of the largest class in $\mathcal{D}_{tr}$. $d_c$ denotes the dimensionality of the low-dimensional representation of $c$. $p_{c,q}^{ref}$ and $p_{c,q}^{(t)}$ denote the distributions of $c$ on the $q$-th low-dimensional coordinate in $\mathcal{D}_{tr}$ and $\mathcal{D}_{tr}^{(t)}$, respectively. $\text{JS-sim}(\cdot,\cdot)$ denotes the distribution similarity defined based on Jensen–Shannon distance (JS). $\text{Rec}_c^{(t)}$ denotes the recall of $c$ on $\mathcal{D}_{\text{val}}$ in round $t$. The predictive entropy $H(\boldsymbol{p}_i)$ of $\boldsymbol{x}_i$ measures the prediction uncertainty, as shown in Eq. (16).

$$H(\boldsymbol{p}_i) = -\sum_{k=1}^{C} p_{ik}\log p_{ik} \tag{16}$$

where $p_{ik}$ denotes the predicted probability that $\boldsymbol{x}_i$ belongs to class $k$, and $k$ is only the index used to enumerate all classes in the predictive distribution. $\mathcal{V}_c$ denotes the index set in $\mathcal{D}_{\text{val}}$ whose labels are $c$. $|\mathcal{V}_c|$ denotes the sample size of $\mathcal{V}_c$. $h_{c,\text{size}}^{(t)}$ denotes the insufficiency of the sample size of $c$. $h_{c,\text{dist}}^{(t)}$ measures the deviation of the current class distribution from the reference distribution. $h_{c,\text{bdry}}^{(t)}$ reflects whether $f_\theta^{(t)}(\cdot)$ sufficiently learns the decision boundary of $c$. $h_{c,\text{unc}}^{(t)}$ measures the average uncertainty of $f_\theta^{(t)}(\cdot)$ on $c$. The components in Eq. (11) are combined with equal weight to compute the information gap score $\gamma_c^{(t)}$ of $c$ in round $t$, as shown in Eq. (17).

$$\gamma_c^{(t)} = \frac{1}{4}\left(h_{c,\text{size}}^{(t)} + h_{c,\text{dist}}^{(t)} + h_{c,\text{bdry}}^{(t)} + h_{c,\text{unc}}^{(t)}\right) \tag{17}$$

$\gamma_c^{(t)}$ is jointly determined by $h_{c,\text{size}}^{(t)}$, $h_{c,\text{dist}}^{(t)}$, $h_{c,\text{bdry}}^{(t)}$, and $h_{c,\text{unc}}^{(t)}$. Equal weights in Eq. (17) avoid favoring any particular augmentation need a priori and minimize the worst-case score perturbation. Specifically, $h_{c,\text{dist}}^{(t)}$ balances beneficial pattern exploration with distribution shift control. Generated data are permitted to deviate moderately from the original empirical distribution to introduce underrepresented variation patterns or boundary samples, while performance on $\mathcal{D}_{\text{val}}$ is used to determine whether these distributional changes benefit $f_\theta^{(t)}(\cdot)$. $\gamma_c^{(t)}$ of all classes are averaged in round $t$, which is denoted as $\Gamma^{(t)}$, as shown in Eq. (18).

$$\Gamma^{(t)} = \frac{1}{C}\sum_{c=1}^{C}\gamma_c^{(t)} \tag{18}$$

Second, $\Gamma^{(t)}$ guides the hierarchical multi-generator collaboration and scheduling in round $t$. It is successively transformed into multiple generator augmentation actions within the class-level, generator-level, and budget-level over multiple rounds. In the class-level, the HMGCS integrates $\Gamma^{(t)}$ and scarcity in $c$ to compute a class-level scheduling distribution $P_{\text{cls}}^{(t)}(c)$, as shown in Eq. (19) and Eq. (20).

$$\lambda^{(t)} = \frac{\frac{1}{C}\sum_{c=1}^{C} h_{c,\text{size}}^{(t)}}{4\Gamma^{(t)} + \varepsilon} \tag{19}$$

$$P_{\text{cls}}^{(t)}(c) = \lambda^{(t)} P_{\text{bal}}^{(t)}(c) + (1-\lambda^{(t)}) P_{\text{info}}^{(t)}(c) \tag{20}$$

where $P_{\text{bal}}^{(t)}(c)$ denotes the balance-driven augmentation probability of $c$ in round $t$. It reflects the sample-size deficit of $c$ relative to the largest class. $P_{\text{info}}^{(t)}(c)$ denotes the information-driven augmentation probability of $c$ in round $t$, which reflects $\gamma_c^{(t)}$. $P_{\text{info}}^{(t)}(c)$ is computed by normalizing $\gamma_c^{(t)}$, and $P_{\text{bal}}^{(t)}(c)$ is computed by normalizing the class deficit $N_{\max}^{ref} - N_c^{(t)}$, where $N_c^{(t)}$ denotes the sample size of $c$ in $\mathcal{D}_{tr}^{(t)}$. $\lambda^{(t)}$ is defined as the proportion of $\Gamma^{(t)}$ contributed by the sample-size deficiency. When $\Gamma^{(t)}$ mainly arises from class imbalance, $P_{\text{cls}}^{(t)}(c)$ is automatically biased toward balance compensation. When $\Gamma^{(t)}$ mainly arises from distribution, boundary, and uncertainty, $P_{\text{cls}}^{(t)}(c)$ is automatically biased toward information-gap compensation. $P_{\text{cls}}^{(t)}(c)$ demonstrates that the HMGCS achieves a balance across classes alongside the multi-round information compensation with $\Gamma^{(t)}$. The classes that should be prioritized for augmentation in $\mathcal{D}_{tr}$ are determined. The class-level decision in round $t$ is made.

In the generator-level, the HMGCS allocates generators according to their capabilities. For $c$ in round $t$, $h_{c,\text{size}}^{(t)}$ and $h_{c,\text{dist}}^{(t)}$ are aggregated into the exploration demand $E_c^{(t)}$, $h_{c,\text{bdry}}^{(t)}$ and $h_{c,\text{unc}}^{(t)}$ are aggregated into the refinement demand $R_c^{(t)}$. By combining $E_c^{(t)}$, $R_c^{(t)}$, $\boldsymbol{S}_{g,c}$, and $\boldsymbol{C}_{g,c}$, the score of each generator is computed, as shown in Eq. (21).

$$\Psi_{g,c}^{(t)} = \mu_0\boldsymbol{S}_{g,c} + \mu_1 E_c^{(t)}\boldsymbol{C}_{g,c} + \mu_2 R_c^{(t)}\boldsymbol{S}_{g,c} \tag{21}$$

where the first term acts as a foundational constraint to ensure that generator selection remains grounded in the real data distribution. The second and third terms adaptively reinforce either exploratory or refining generators, guided by the class-level decision in round $t$. $\Psi_{g,c}^{(t)}$ enables generators with high $\boldsymbol{C}_{g,c}$ to prioritize coverage expansion and distribution completion, while generators with high $\boldsymbol{S}_{g,c}$ prioritize boundary refinement and uncertainty reduction. $\mu_0$, $\mu_1$, and $\mu_2$ are derived from the normalization of $E_c^{(t)}$ and $R_c^{(t)}$, enabling dynamic adaptation of weights to the class-level decision. Greater exploration demand favors generators with high $\boldsymbol{C}_{g,c}$, whereas greater refinement demand favors those with high $\boldsymbol{S}_{g,c}$. Generator allocation therefore adapts to the evolving information-gap composition of each class across augmentation rounds. The generator distribution $P_{\text{gen}}^{(t)}(g \mid c)$ in $c$ is then normalized from $\Psi_{g,c}^{(t)}$, as shown in Eq. (22).

$$P_{\text{gen}}^{(t)}(g \mid c) = \frac{\Psi_{g,c}^{(t)}}{\sum_{g'=1}^{G} \Psi_{g',c}^{(t)} + \varepsilon} \tag{22}$$

where $g'$ denotes the summation index across generators. The generator-level decision is determined by the mapping between the information gap and the multiple generator capabilities.

In the budget-level, the class-level and generator-level decisions are further converted into generated sample allocations. In each round, $\Gamma^{(t)}$ decomposed according to $P_{\text{cls}}^{(t)}(c)$ in the class-level decisions and $P_{\text{gen}}^{(t)}(g \mid c)$ in the generator decision, as shown in Eq. (23).

$$n_c^{(t)} \leftarrow B^{(t)} P_{\text{cls}}^{(t)}(c), \qquad n_{g,c}^{(t)} \leftarrow n_c^{(t)} P_{\text{gen}}^{(t)}(g \mid c) \tag{23}$$

where $n_c^{(t)}$ denotes the total budget of augmented data allocated to $c$ in round $t$, and $n_{g,c}^{(t)}$ denotes the budget allocated to $g$ in $c$. Therefore, $n_c^{(t)}$ and $n_{g,c}^{(t)}$ collaborate and schedule the multiple generators. The budget-level decision in round $t$ is made. In each round, the HMGCS provides generated sample allocations for each class through the class-level decision, the generator-level decision, and the budget-level decision. The algorithm flow of the HMGCS is described in Algorithm 2.

**Algorithm 2** HMGCS

**Require:** $\mathcal{D}_{\text{tr}}, \mathcal{D}_{\text{tr}}^{(t)}, \mathcal{D}_{\text{val}}, f_\theta^{(t)}(\cdot), \boldsymbol{A}$
**Ensure:** $n_{g,c}^{(t)}$

```
for x_i ∈ D_val do
    p_i ← f_θ^(t)(x_i) using Eq. (10)
end for
for c ∈ Y do
    h_c^(t) ← D_tr, D_tr^(t), {p_i}_{i∈V_c} using Eqs. (11)–(16)
    γ_c^(t) ← h_c^(t) using Eq. (17)
end for
Γ^(t) ← {γ_c^(t)}_{c=1}^C using Eq. (18)
for c ∈ Y do
    P_cls^(t)(c) ← γ_c^(t), N_c^(t), N_max^ref using Eqs. (19) and (20)
    for g ∈ {g_n}_{n=1}^G do
        Ψ_{g,c}^(t) ← h_c^(t), A_{g,c} using Eq. (21)
    end for
    P_gen^(t)(g | c) ← {Ψ_{g,c}^(t)}_{g=1}^G using Eq. (22)
    n_c^(t), n_{g,c}^(t) ← P_cls^(t)(c), P_gen^(t)(g | c), B^(t) using Eq. (23)
end for
return n_{g,c}^(t)
```

### C. IGMP-EC

The IGMP-EC is proposed to evaluate and perform accept-or-reject decisions on candidate augmentation over multiple rounds. The IGMP-EC and the HMGCS jointly form a closed loop over multiple rounds. Data augmentation terminates automatically when information-gap reduction reaches saturation and downstream model improvement plateaus. The IGMP-EC includes an information gap–model performance joint evaluation metric (IGMP-E) and an information gap–model performance joint closed-loop method (IGMP-C).

The IGMP-E is proposed to jointly evaluate model performance improvement and information gap reduction, as shown in Eq. (24).

$$\begin{aligned}\mathcal{J}^{(t)} = \eta_1 \,\text{MacroF1}^{(t)} + \eta_2 \,\text{Acc}^{(t)} + \eta_3 \,\text{MacroRec}^{(t)} \\ + \eta_4 \,\text{MacroPre}^{(t)} - \eta_5 \,\Gamma^{(t)}\end{aligned} \tag{24}$$

where $\text{MacroF1}^{(t)}$, $\text{Acc}^{(t)}$, $\text{MacroRec}^{(t)}$, and $\text{MacroPre}^{(t)}$ denote the Macro-F1, accuracy, Macro-Recall, and Macro-Precision on $\mathcal{D}_{val}$ in round $t$, respectively. $\eta_1$ to $\eta_5$ are non-negative weights. $\Gamma$ is the initial value of $\Gamma^{(t)}$, and the initial value of $\mathcal{J}^{(t)}$ is $\mathcal{J}$. $\Gamma$ and $\mathcal{J}$ are evaluated on $\mathcal{D}_{val}$. $\text{MacroF1}^{(t)}$, $\text{Acc}^{(t)}$, $\text{MacroRec}^{(t)}$, and $\text{MacroPre}^{(t)}$ on $\mathcal{D}_{val}$ also jointly reflect the validation utility of the data produced by the generators. This utility is also taken into account in the decisions made in each round. It is worthy of note that although recall appears in both Eqs. (14) and (24), it supports class-specific diagnosis of augmentation needs in the former and overall evaluation of candidate augmentations in the latter.

In the IGMP-C, a closed-loop method of launch injection followed by refinement adjustment is proposed. First, the launch injection serves as round 0 with the sample size $B^{(0)}$. An initial augmentation explores whether the data augmentation can improve $\mathcal{J}^{(t)}$. $B^{(0)}$ is shown in Eq. (25).

$$B^{(0)} = \text{round}\big(\kappa\,(B_{\min} + (B_{\max} - B_{\min})\Gamma)\big) \tag{25}$$

where $B_{\min}$ and $B_{\max}$ denote the lower and upper bounds of the budget, respectively. $\kappa$ denotes the amplification factor for rapidly compensating for the significant information gap. $\text{round}(\cdot)$ denotes the rounding operator. $B^{(0)}$ is first allocated into $n_c^{(0)}$ according to $P_{\text{cls}}^{(0)}(c)$, and then into $n_{g,c}^{(0)}$ according to $P_{\text{gen}}^{(0)}(g \mid c)$. $n_c^{(0)}$ denotes the sample size allocated to $c$, and $n_{g,c}^{(0)}$ denotes the sample size allocated to $g$ under $c$ in round 0. An improved $\mathcal{J}^{(0)}$ in the launch injection indicates that the model performance improves and $\Gamma^{(t)}$ decreases. Data augmentation is effective. $B^{(0)}$ is the starting point of the closed loop, and the augmentation directly enters the refinement adjustment. Otherwise, data augmentation is stopped.

Second, in the refinement adjustment, the initial budget of the refinement adjustment $B^{(1)}$ is updated by Eq. (25), where $\kappa$ is removed to avoid over-augmentation and enable finer-grained adjustment, thereby improving the stability of the closed loop. For round $t$ ($t \geq 2$), the sample size of the candidate augmentation $\Delta\mathcal{D}^{(t)}$ is defined in Eq. (26).

$$B^{(t+1)} = \text{round}\big(B^{(t)} \cdot \rho_r^{\,t}\big) \tag{26}$$

where $\rho_r$ denotes the cross-round shrinkage factor. In round $t$, the class-wise sample size and the generator-wise sample size are determined by $P_{\text{cls}}^{(t)}(c)$ and $P_{\text{gen}}^{(t)}(g \mid c)$ in the HMGCS, respectively. Whether $\Delta\mathcal{D}^{(t)}$ is accepted and whether $B^{(t)}$ is updated both depend on $\mathcal{J}^{(t)}$ and $\Gamma^{(t)}$. If $\mathcal{J}^{(t)}$ increases and $\Gamma^{(t)}$ decreases, the exploration of the augmented sample size in subsequent rounds is shrunk by $\rho_r$. $\Delta\mathcal{D}^{(t)}$ is accepted. If $\mathcal{J}^{(t)}$ decreases or $\Gamma^{(t)}$ increases, the hierarchical multiple generators are repeatedly invoked within the same round to generate data with the same size $B^{(t)}$. The same size reduces the influence of randomness on $\mathcal{J}^{(t)}$. $\Delta\mathcal{D}^{(t)}$ is rejected. If $\mathcal{J}^{(t,\ell)}$ still does not improve after $s$ attempts ($\forall \ell \in \{1, \ldots, s\}$), the IGMP-C enters the next round $t+1$ with $B^{(t+1)}$ scaled by $\rho_r$. The total budget is initialized from $\Gamma$ using Eq. (25) and subsequently updated in Eq. (26). It is worth noting that even if $\Delta\mathcal{D}^{(t)}$ is rejected in round $t$, $B^{(t+1)}$ is still updated using Eq. (26). A smaller $\Delta\mathcal{D}^{(t+1)}$ is then augmented to re-explore the effective data space for augmentation.

However, the growth of $\mathcal{J}^{(t)}$ is not unbounded. The stopping conditions of the IGMP-C include two cases. One is that $\Gamma^{(t)}$ falls below 5% of $\Gamma$. In practice, the information gap is generally difficult to eliminate completely through data augmentation. Therefore, a residual gap of 5% of the initial information gap is allowed, thereby relaxing the idealized requirement of complete gap elimination and avoiding continued sample generation solely to achieve a zero gap. The other is that no acceptable candidate augmentation appears for multiple consecutive rounds $r_{max}$. Therefore, the IGMP-EC augments only those $\Delta\mathcal{D}^{(t)}$ that both improve the performance of $f_\theta^{(t)}(\cdot)$ and reduce $\Gamma^{(t)}$ over multiple rounds. $r_{\text{rej}}$ serves as a counter to track the number of consecutive rounds in which model updates are rejected. The IGMP-EC evaluates and performs accept-or-reject decisions on candidate augmentation over multiple rounds, as described in Algorithm 3.

**Algorithm 3** IGMP-EC

**Require:** $n_{g,c}^{(t)}, \mathcal{D}_{\text{tr}}, \mathcal{D}_{\text{val}}, f_\theta(\cdot), \Gamma, \mathcal{J}$
**Ensure:** $\mathcal{D}_{\text{tr}}^{(t)}$, $f_\theta^{(t)}(\cdot)$

**IGMP-C:** $B^{(0)} \leftarrow \Gamma$, $B_{\min}$, $B_{\max}$ using Eq. (25)
$\Delta\mathcal{D}^{(0)} \leftarrow n_{g,c}^{(0)}(\text{HMGCS})$
$\mathcal{D}_{\text{tr}}^{(0)} \leftarrow \mathcal{D}_{\text{tr}}$, $\Delta\mathcal{D}^{(0)}$
$f_\theta^{(0)}(\cdot)$, $\Gamma^{(0)} \leftarrow \mathcal{D}_{\text{tr}}^{(0)}$, $\mathcal{D}_{\text{val}}$
**IGMP-E:** $\mathcal{J}^{(0)} \leftarrow f_\theta^{(0)}(\cdot)$, $\Gamma^{(0)}$ using Eq. (24)
$\Delta\mathcal{J}^{(0)} \leftarrow \mathcal{J}^{(0)}$, $\mathcal{J}$
**if** $\Delta\mathcal{J}^{(0)} \leq 0$ **or** $\Gamma^{(0)} \geq \Gamma$ **then**
  **return** $\mathcal{D}_{\text{tr}}$, $f_\theta(\cdot)$
**end if**
$B^{(1)} \leftarrow \Gamma^{(0)}$, $B_{\min}$, $B_{\max}$ using Eq. (25) without $\kappa$
$r_{\text{rej}} \leftarrow 0$
**for** $t = 1, 2, 3, \ldots$ **do**
  $\Delta\mathcal{D}^{(t)} \leftarrow n_{g,c}^{(t)}(\text{HMGCS})$
  $f_\theta^{(t)}(\cdot)$, $\Gamma^{(t)} \leftarrow \mathcal{D}_{\text{tr}}^{(t-1)}$, $\Delta\mathcal{D}^{(t)}$, $\mathcal{D}_{\text{val}}$
  **IGMP-E:** $\mathcal{J}^{(t)} \leftarrow f_\theta^{(t)}(\cdot)$, $\Gamma^{(t)}$ using Eq. (24)
  $\Delta\mathcal{J}^{(t)} \leftarrow \mathcal{J}^{(t)}$, $\mathcal{J}^{(t-1)}$
  **if** $\Delta\mathcal{J}^{(t)} > 0$ **and** $\Gamma^{(t)} < \Gamma^{(t-1)}$ **then**
    $\mathcal{D}_{\text{tr}}^{(t)} \leftarrow \mathcal{D}_{\text{tr}}^{(t-1)}$, $\Delta\mathcal{D}^{(t)}$
    $B^{(t+1)} \leftarrow B^{(t)}$, $\rho_r$ using Eq. (26)
    $r_{\text{rej}} \leftarrow 0$
  **else**
    **for** $\ell = 1, 2, \ldots, s$ **do**
      $\Delta\mathcal{D}^{(t)} \leftarrow n_{g,c}^{(t)}(\text{HMGCS})$
      $f_\theta^{(t,\ell)}(\cdot)$, $\Gamma^{(t,\ell)} \leftarrow \mathcal{D}_{\text{tr}}^{(t-1)}$, $\Delta\mathcal{D}^{(t)}$, $\mathcal{D}_{\text{val}}$
      **IGMP-E:** $\mathcal{J}^{(t,\ell)} \leftarrow f_\theta^{(t,\ell)}(\cdot)$, $\Gamma^{(t,\ell)}$ using Eq. (24)
      $\Delta\mathcal{J}^{(t,\ell)} \leftarrow \mathcal{J}^{(t,\ell)}$, $\mathcal{J}^{(t-1)}$
      **if** $\Delta\mathcal{J}^{(t,\ell)} > 0$ **and** $\Gamma^{(t,\ell)} < \Gamma^{(t-1)}$ **then**
        $\mathcal{D}_{\text{tr}}^{(t)} \leftarrow \mathcal{D}_{\text{tr}}^{(t-1)}$, $\Delta\mathcal{D}^{(t)}$
        $\mathcal{J}^{(t)}$, $\Gamma^{(t)}$, $f_\theta^{(t)}(\cdot) \leftarrow \mathcal{J}^{(t,\ell)}$, $\Gamma^{(t,\ell)}$, $f_\theta^{(t,\ell)}(\cdot)$
        $B^{(t+1)} \leftarrow B^{(t)}$, $\rho_r$ using Eq. (26)
        $r_{\text{rej}} \leftarrow 0$
        **break**
      **end if**
    **end for**
    **if** $\Delta\mathcal{J}^{(t,\ell)} \leq 0$ **or** $\Gamma^{(t,\ell)} \geq \Gamma^{(t-1)}$, $\forall \ell \in \{1, \ldots, s\}$ **then**
      $\mathcal{D}_{\text{tr}}^{(t)} \leftarrow \mathcal{D}_{\text{tr}}^{(t-1)}$
      $\mathcal{J}^{(t)}$, $\Gamma^{(t)}$, $f_\theta^{(t)}(\cdot) \leftarrow \mathcal{J}^{(t-1)}$, $\Gamma^{(t-1)}$, $f_\theta^{(t-1)}(\cdot)$
      $B^{(t+1)} \leftarrow B^{(t)}$, $\rho_r$ using Eq. (26)
      $r_{\text{rej}} \leftarrow r_{\text{rej}} + 1$
    **end if**
  **end if**
  **if** $\Gamma^{(t)} \leq 0.05\,\Gamma$ **or** $r_{\text{rej}} \geq r_{\max}$ **then**
    **break**
  **end if**
**end for**
**return** $\mathcal{D}_{\text{tr}}^{(t)}$, $f_\theta^{(t)}(\cdot)$

## IV. Experiments

A micro weather station integrating multiple IoT sensors served as an integrated sensing node and edge data source in industrial, agricultural, and environmental monitoring. Therefore, we conducted comprehensive experiments on the micro weather station with IoT sensors in WSNs to validate the IGADA-IoT. The micro weather station was named “Fuxi”. Experiments consisted of five parts: experimental environment and dataset preparation, generation evaluation and capability modeling in the GCM, exploration of data augmentation with the IGADA-IoT, ablation study, and real-world deployments.

### A. Experimental Environment and Dataset Preparation

*1) Experimental Environment:* “Fuxi” continuously acquired multi-source sensing data in WSNs, as shown in Fig. 2(a). It integrated 6 IoT sensors, including light, pressure, temperature, humidity, soil temperature, and soil humidity. The data acquired by each sensor served as one feature. The acquired data were represented as multivariate time series and were transformed into time windows to describe the local environmental variation state in $T$. $\mathcal{Y}$ contained 3 discrete classes: 0, 1, and 2. Class 0 indicated that environments were stable and the sampling frequency was reduced, class 1 indicated normal data acquisition, and class 2 indicated that environments were unstable and the sampling frequency was increased. The sampling frequency for the normal data acquisition of “Fuxi” was once every 4 minutes. Class 0 indicated that the sampling frequency should be reduced to once every 8 minutes to save energy. Class 1 indicated that no adjustment of the sampling frequency was required. Class 2 indicated that the sampling frequency should be increased to once every 2 minutes to improve the information density of the acquired data. Predicting a class associated with a shorter sampling interval than the reference label increased acquisition counts and the associated sampling energy consumption. Conversely, predicting a longer interval could leave rapid environmental changes insufficiently captured, potentially causing information loss. Environmental changes exhibited short-term continuity and hysteresis, $T$ was set to 15 to avoid incorrect decisions caused by relying only on instantaneous observations.

*2) Dataset Preparation:* The acquired data were cleaned, aligned, split by date, and transformed into time windows within each date. No overlapping windows from the same date appeared in different subsets. Next, the dataset was divided into $\mathcal{D}_{tr}$, $\mathcal{D}_{val}$, and $\mathcal{D}_{test}$ in a ratio of 7:2:1, as shown in Fig. 2(b). Sample labels were obtained through manual annotation of a small subset, clustering, and final manual verification. Density-based spatial clustering of applications with noise (DBSCAN) was used for its ability to identify nonspherical clusters and flag isolated observations in multivariate sensor data. To avoid information leakage, data augmentation was performed only on $\mathcal{D}_{tr}$. In $\mathcal{D}_{tr}$ and $\mathcal{D}_{val}$, there were 259 samples in class 0, 503 samples in class 1, and 228 samples in class 2. $\mathcal{D}_{tr}$ and $\mathcal{D}_{val}$ together contained a total of 990 samples, with 770 samples in $\mathcal{D}_{tr}$ and 220 in $\mathcal{D}_{val}$. The sample size of $\mathcal{D}_{tr}$ was limited and the minority classes were particularly scarce, making it a typical class-imbalanced and scale-limited dataset. In the IGADA-IoT, $\mathcal{D}_{tr}$ was used for data augmentation in each round, and $\mathcal{D}_{val}$ was used to validate the improvement in model performance and the reduction in the information gap. It was noteworthy that $\mathcal{D}_{test}$ did not participate in any augmentation or decision and was used only for external observation. The fairness and objectivity of the evaluation were ensured.

*3) Experiment Setup:* The NVIDIA GeForce RTX 3090 was used for training the generators and data augmentation. The 13th Gen Intel Core i7-13700F CPU was used for the IGADA-IoT and training the downstream models. Furthermore, data augmentation ran offline, with resource costs separate from those of downstream training, inference, and deployment, and was not constrained by edge-device resources.

### *B. Generation Evaluation and Capability Modeling in GCM*

In the IGADA-IoT, the generators adopted a "primary–auxiliary" hierarchical architecture. The primary generator was a generative model. Its high diversity and strong distribution modeling capability were used for exploratory information compensation. The auxiliary generators were generation strategies. Their high similarity and local stability were used for fine-grained information refinement. In the "primary–auxiliary" hierarchical architecture, multiple generators complemented and balanced the expansion of coverage and the fitting of the data space to the real distribution. An excessive number of primary generators would introduce redundant exploration directions and accumulated distribution shift, thereby reducing the stability of information compensation. A single primary generator was more suitable for concentrating the major information expansion. The diversification of auxiliary generators could improve the directional diversity and class adaptability of local refinement. Boundary and fine-grained details were complemented more sufficiently. Therefore, the

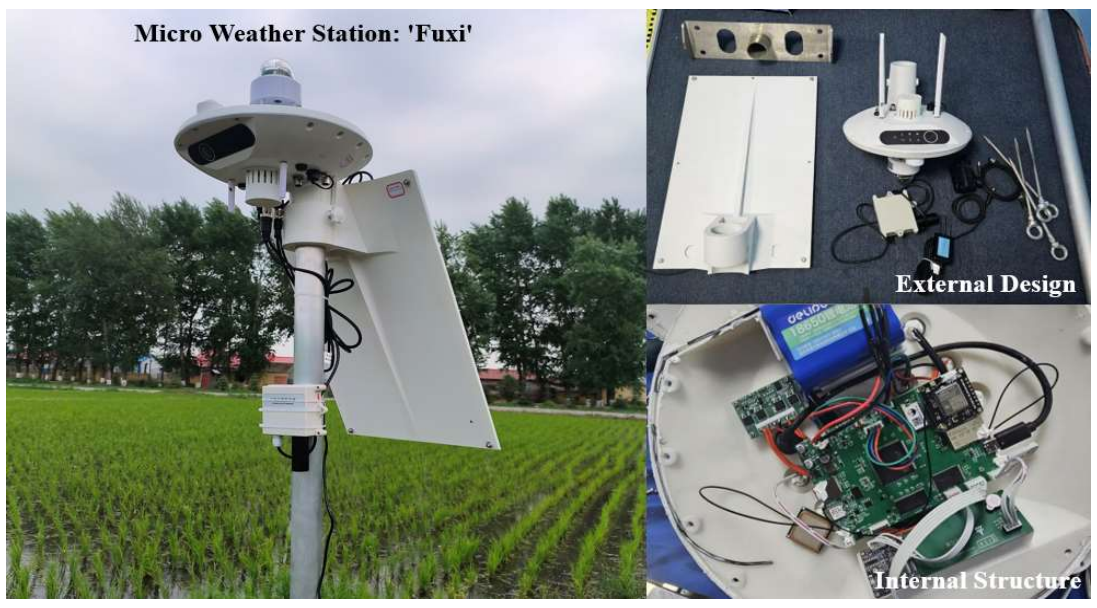

(a) Micro weather station "Fuxi", including its external design and internal structure.

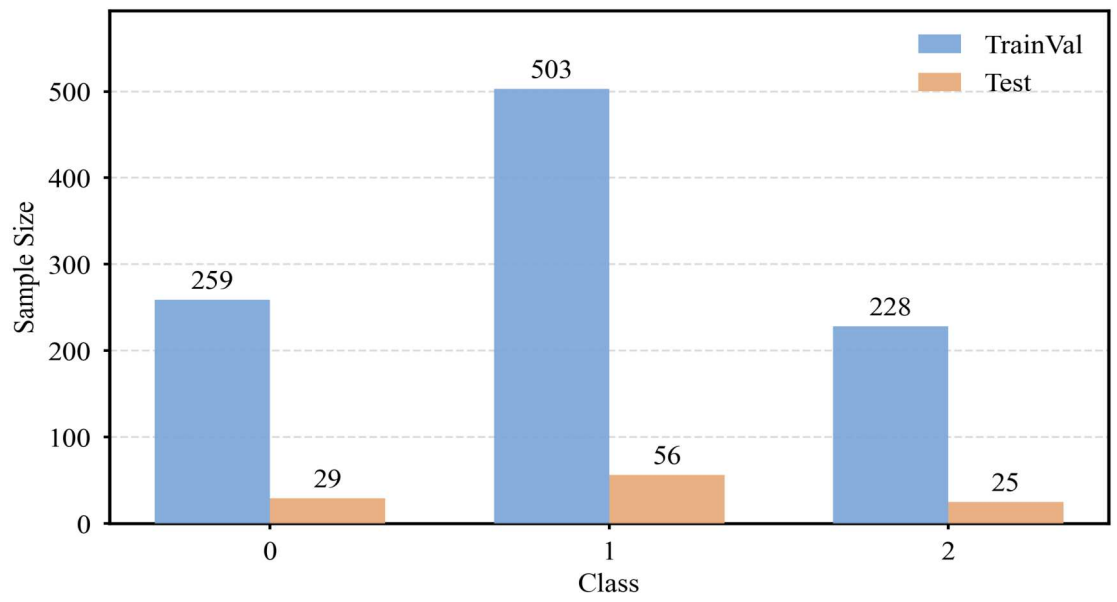

(b) Sample size and class distribution in IoT sensor dataset.

Fig. 2. Micro weather station "Fuxi" and IoT sensor dataset.

primary generator adopted the ImagenTime [47], which was a state-of-the-art diffusion model for time-series generation. The auxiliary generators adopted the advanced generation strategies, including RIM [34], DTW [35], and TSW [36] for time-series generation.

*1) Qualitative Evaluation:* The qualitative evaluation aimed to visually verify whether different generators could preserve the sequential patterns of the real data and the coupling relationships among multiple variables, thereby providing interpretable support for generator capability modeling. In Fig. 3, the mean values of real data and generated data, and their fluctuation ranges were compared for the 3 classes across the 6 features. The fitting performance of the generated data was illustrated in terms of the trends, class discriminability, and local variation details. The comparisons in Fig. 3 showed that all generators preserved the distribution of the real data well. The ImagenTime followed the real trajectories more adequately on features with more complex variations, such as light, pressure, and temperature, and in particular achieved better coverage of the fluctuations in class 2. In contrast, the RIM, the DTW, and the TSW exhibited better local fidelity and stability on relatively smooth features, such as humidity, soil temperature, and soil humidity, but their ability to extend highly volatile patterns was relatively weaker. The qualitative evaluation demonstrated the clear complementarity among different generators, and it also provided intuitive evidence for the "primary–auxiliary" hierarchical architecture of generators in the IGADA-IoT.

*2) Quantitative Evaluation and Generator Capability Modeling:* The quantitative evaluation not only showed the quality of the generated data, but also quantified generator capability for modeling. In $\mathcal{D}_{tr}$, $\boldsymbol{C}_{g,c}$ and $\boldsymbol{S}_{g,c}$ of the generated data were computed for all classes. Moreover, the general metrics in the time-series generation task (denoted as G. M.), predictive score (Pred) and discriminative score (Disc), as well as Kullback-Leibler divergence (KL) and JS, were used to evaluate the generated data. G. M. denoted the universal metrics used to evaluate generators and their generated data. More details of Pred and Disc can be found in [47]. To reduce randomness, the mean ± standard deviation of Pred and Disc were shown as the experimental result. $\boldsymbol{C}_{g,c}$, $\boldsymbol{S}_{g,c}$, KL, and JS were derived from the mean of multiple data groups. The quantitative evaluation results were shown in Table I.

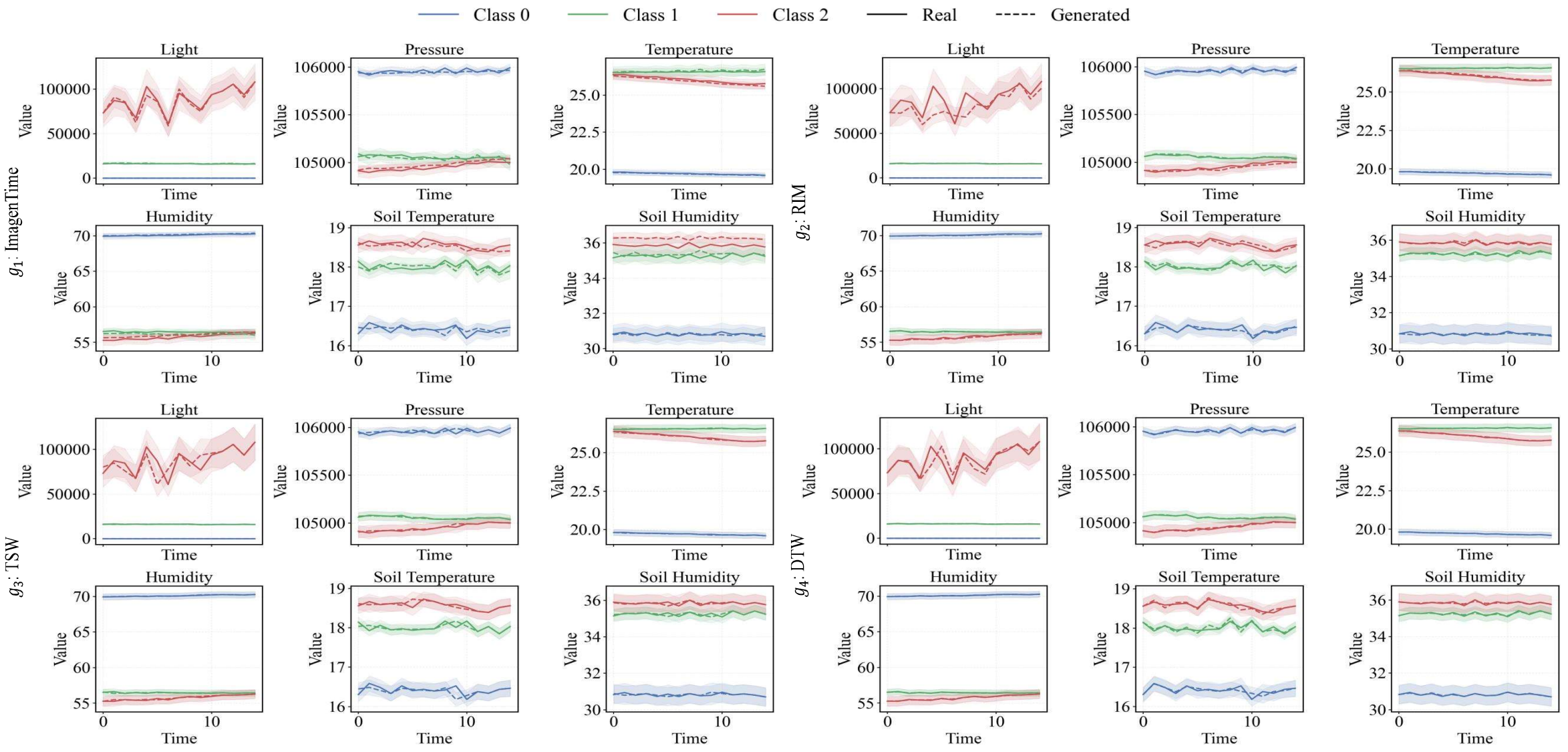

Fig. 3. Comparison between real data and generated data in qualitative evaluation of generators, including ImagenTime, RIM, TSW, and DTW.

TABLE I
QUANTITATIVE EVALUATION OF GENERATORS IN ALL CLASSES.

| | Class | Generator | | | |
|---|---|---|---|---|---|
| | | ImagenTime | RIM | TSW | DTW |
| $\boldsymbol{C}_{g,c}$ | 0 | 0.999 | 0.552 | 0.312 | 0.449 |
| | 1 | 1 | 0 | 0 | 0 |
| | 2 | 0.998 | 0.249 | 0 | 0 |
| $\boldsymbol{S}_{g,c}$ | 0 | 0.001 | 0.448 | 0.688 | 0.551 |
| | 1 | 0 | 1 | 1 | 1 |
| | 2 | 0.002 | 0.751 | 1 | 1 |
| G. M. | Pred (↓) | .219±.010 | .336±.012 | .331±.018 | .335±.018 |
| | Disc (↓) | .003±.002 | .001±.000 | .002±.001 | .001±.001 |
| | KL (↓) | 1.608 | 1.136 | 1.140 | 1.149 |
| | JS (↓) | 0.248 | 0.212 | 0.146 | 0.152 |

TABLE II
STRUCTURES AND PARAMETERS OF DOWNSTREAM MODELS.

| Model | Type | Parameters |
|---|---|---|
| LR | Linear | $C = 1.0$, L2 regularization, multinomial loss, balanced class weights |
| RF | Tree ensemble | 100 trees, max depth = 5, min leaf = 2, max features = sqrt |
| SVC | Kernel-based | kernel = "rbf", $C = 10.0$, gamma = scale |
| ANN | MLP | Hidden dimensions: [512, 256, 512] |
| 1D-CNN | Convolutional | Channels: [128, 256], kernels: [5, 3], classifier hidden = 256 |
| GRU | Recurrent | Hidden size = 64, layer = 2, classifier hidden = 128 |
| LSTM | Recurrent | Hidden size = 64, layer = 2, classifier hidden = 128 |
| Transformer | Attention-based | $d_{\text{model}} = 64$, head = 8, encoder layer = 3, FFN dim = 128 |

Table I showed that $\boldsymbol{C}_{g,c}$ of the ImagenTime was close to 1 for all classes, while its similarity was close to 0. The results indicated that the ImagenTime provided the strongest information expansion capability. The ImagenTime therefore served as the primary generator for large-scale information-gap compensation. In contrast, the RIM still retained a certain degree of information contribution in classes 0 and 2, while its similarity reached 1 in class 1. $\boldsymbol{C}_{g,c}$ of the TSW and the DTW were generally weaker, but they exhibited higher similarity in classes 1 and 2. The generation strategies were more effective at preserving the local structure of real data and the class-conditional stability. They therefore served as auxiliary generators for boundary and detail refinement. The results in Table I were also consistent with this pattern. Pairing for augmentation strategies preserved sample-level dependence and could yield higher mutual information estimates, whereas independent within-class pairing for generative models typically yielded estimates near 0. Furthermore, the general metrics and divergence metrics were further used for verification. The ImagenTime achieved the best Pred, indicating that its generated data had stronger utility for the downstream task. The generation strategies achieved lower KL and JS divergences overall, and their Disc also remained at low levels. Their generated data were closer to the real distribution and more stable. Therefore, the experimental results provided quantitative evidence for the "primary–auxiliary" hierarchical architecture of multiple generators in the IGADA-IoT.

For each class, $\boldsymbol{C}_{g,c}$ and $\boldsymbol{S}_{g,c}$ were used to form $\boldsymbol{A}_{g,c}$, which was further tensorized in $\boldsymbol{A}$ in the IGADA-IoT.

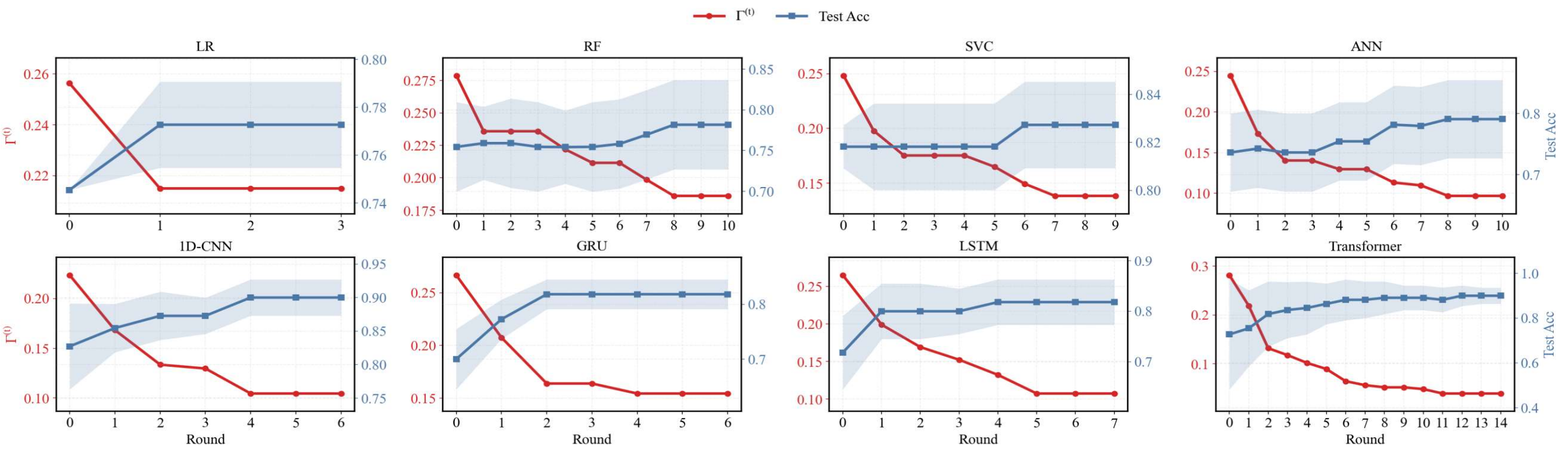

Fig. 4. $\Gamma^{(t)}$ and Test Acc over multiple rounds.

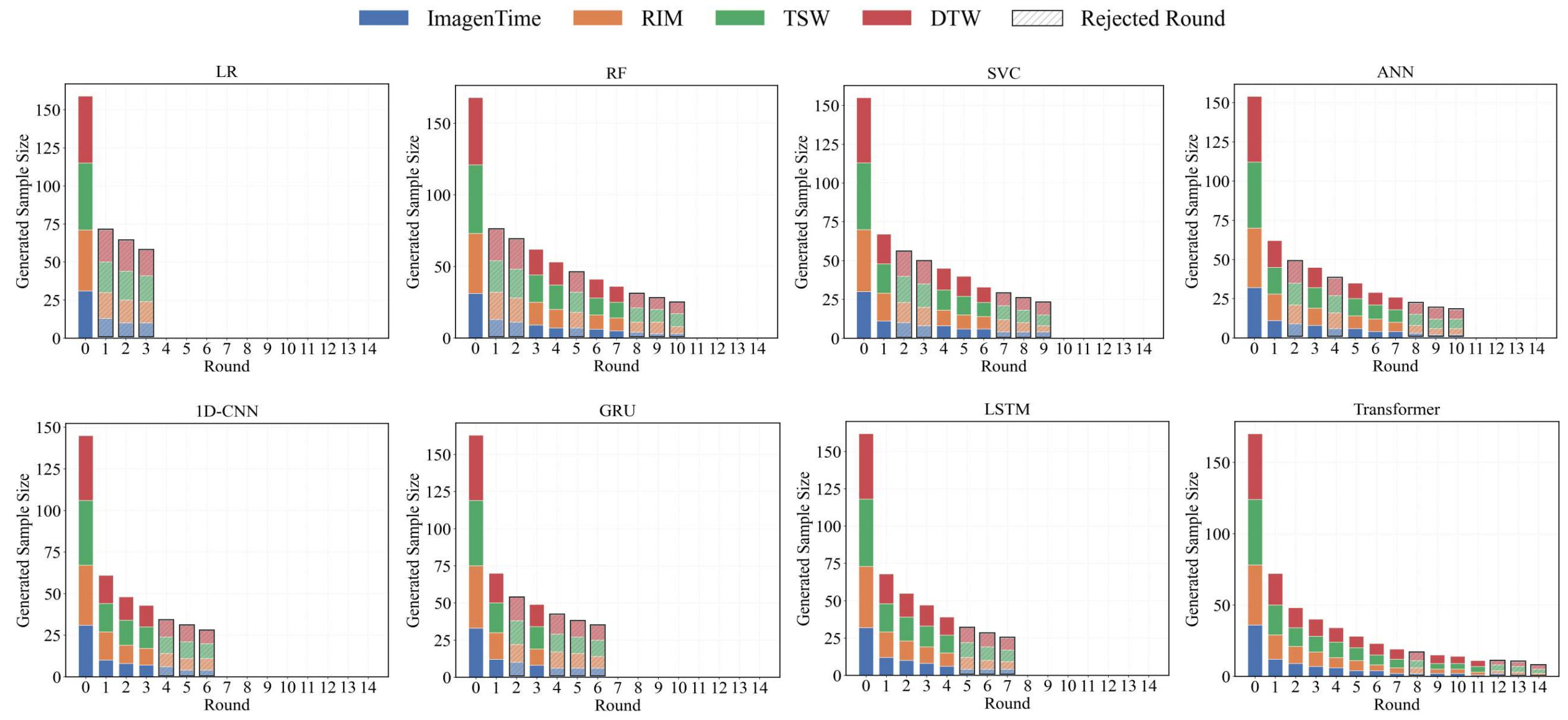

Fig. 5. Generated sample size, budget allocations, and accept/reject decisions over multiple rounds (dark-gray outlines indicate rejected rounds).

### *C. Exploration of Data Augmentation with IGADA-IoT*

*1) Downstream Model Structure and Parameters:* To ensure that the experiments were not biased toward a specific classifier, 8 downstream models with different levels of complexity were used, including LR, RF, SVC, ANN, 1D-CNN, GRU, LSTM, and Transformer [20], [21], [26], [27], [28], [29]. The structures and parameters of the downstream models were shown in Table II. The 8 models covered commonly used model families for energy optimization, ranging from machine learning models to deep learning models. The hyperparameters of the models were set via grid search, and K-fold cross-validation was employed. To reduce randomness, random seeds 42–51 were used. The following experiments further illustrated the automatic data augmentation in the IGADA-IoT.

*2) HMGCS and IGMP-EC for Exploration over Multiple Rounds:* To validate the HMGCS and the IGMP-EC in exploration, the experiments in this subsection were conducted. $B_{\min}$ and $B_{\max}$ were set to 30 and 300 (rounding half the training-set size down to the nearest multiple of 100), respectively. Given the size of $\mathcal{D}_{\text{tr}}$, $B_{\min}$ ensured that the number of generated samples was not excessively small, allowing subsequent generated samples to provide a substantial quantitative supplement to $\mathcal{D}_{\text{tr}}$. [51], [52]. $\kappa$ was set to 1.6. $\rho_r$ was set to 0.9 to ensure stable exploration. $s$ and $r_{\max}$ were both fixed at 3. $\eta_1$ to $\eta_5$ were set to 0.5, 0.3, 0.1, 0.1, and 0.2 through preliminary validation tuning, where higher weights of $\mathrm{MacroF1}^{(t)}$, $\mathrm{Acc}^{(t)}$, $\Gamma^{(t)}$ indicated their dominant roles in the guidance. The PCA dimension was set equal to the number of IoT sensors to balance information preservation and computational cost. These hyperparameters determined the generated sample size. To verify the robustness of the hyperparameters, all subsequent comparative methods were evaluated under the same sample size. The accuracy on $\mathcal{D}_{\text{test}}$ over multiple rounds (denoted as Test Acc) was shown post hoc only as an objective and fair external observation of the sampling-frequency decision model performance over multiple rounds in data augmentation, rather than being used as optimization feedback. Test Acc was not used for model training, collaboration and scheduling, accept-or-reject decisions, or stopping conditions in the IGADA-IoT. Fig. 4 showed $\Gamma^{(t)}$ and Test Acc. $\Gamma^{(t)}$ and Test Acc described the multi-round automatic data augmentation from two complementary perspectives: external model performance and internal information deficiency. Fig. 5 showed the generated sample size, the budget allocations, and the accept-or-reject decisions over multiple rounds in detail. Moreover, the experimental results in Table III quantitatively showed $\Delta\mathcal{J}^t$, $\Delta\Gamma^{(t)}$ (the variation in $\Gamma^{(t)}$), and the accept-or-reject decisions of the IGADA-IoT in multi-round automatic data augmentation. "Accept" denoted the accept-or-reject decision, with 1 denoting acceptance and 0 denoting rejection. Table IV showed the comprehensive results of the IGADA-IoT on the 8 downstream models.

TABLE III
AUTOMATIC DATA AUGMENTATION IN IGADA-IOT.

| Model | Metric | Round | | | | | | | | | | | | | | |
|---|---|---|---|---|---|---|---|---|---|---|---|---|---|---|---|---|
| | | R0 | R1 | R2 | R3 | R4 | R5 | R6 | R7 | R8 | R9 | R10 | R11 | R12 | R13 | R14 |
| LR | $\Delta\mathcal{J}^t$ | 0.030 | -0.016 | -0.007 | -0.007 | – | – | – | – | – | – | – | – | – | – | – |
| | $\Delta\Gamma^{(t)}$ | 0.041 | 0.010 | 0.013 | 0.011 | – | – | – | – | – | – | – | – | – | – | – |
| | Accept | 1 | 0 | 0 | 0 | – | – | – | – | – | – | – | – | – | – | – |
| RF | $\Delta\mathcal{J}^t$ | 0.016 | -0.010 | -0.007 | 0.005 | 0.005 | -0.003 | 0.021 | 0.020 | -0.006 | -0.008 | -0.043 | – | – | – | – |
| | $\Delta\Gamma^{(t)}$ | 0.043 | 0.013 | 0.011 | 0.014 | 0.010 | 0.009 | 0.013 | 0.012 | 0.006 | 0.004 | -0.004 | – | – | – | – |
| | Accept | 1 | 0 | 0 | 1 | 1 | 0 | 1 | 1 | 0 | 0 | 0 | – | – | – | – |
| SVC | $\Delta\mathcal{J}^t$ | 0.026 | 0.011 | -0.006 | -0.008 | 0.005 | 0.013 | 0.013 | -0.015 | -0.004 | -0.016 | – | – | – | – | – |
| | $\Delta\Gamma^{(t)}$ | 0.050 | 0.023 | 0.012 | 0.012 | 0.010 | 0.016 | 0.011 | 0.004 | 0.004 | 0.002 | – | – | – | – | – |
| | Accept | 1 | 1 | 0 | 0 | 1 | 1 | 1 | 0 | 0 | 0 | – | – | – | – | – |
| ANN | $\Delta\mathcal{J}^t$ | 0.131 | 0.016 | -0.030 | 0.003 | -0.008 | 0.006 | 0.004 | 0.026 | -0.001 | -0.039 | -0.006 | – | – | – | – |
| | $\Delta\Gamma^{(t)}$ | 0.071 | 0.033 | -0.007 | 0.011 | 0.003 | 0.016 | 0.003 | 0.013 | 0.014 | -0.004 | 0.012 | – | – | – | – |
| | Accept | 1 | 1 | 0 | 1 | 0 | 1 | 1 | 1 | 0 | 0 | 0 | – | – | – | – |
| 1D-CNN | $\Delta\mathcal{J}^t$ | 0.032 | 0.060 | 0.013 | 0.024 | -0.013 | -0.013 | -0.025 | – | – | – | – | – | – | – | – |
| | $\Delta\Gamma^{(t)}$ | 0.055 | 0.035 | 0.004 | 0.025 | 0.007 | -0.001 | 0.001 | – | – | – | – | – | – | – | – |
| | Accept | 1 | 1 | 1 | 1 | 0 | 0 | 0 | – | – | – | – | – | – | – | – |
| GRU | $\Delta\mathcal{J}^t$ | 0.129 | 0.049 | -0.006 | 0.020 | -0.001 | -0.002 | -0.003 | – | – | – | – | – | – | – | – |
| | $\Delta\Gamma^{(t)}$ | 0.059 | 0.044 | 0.011 | 0.009 | 0.026 | 0.012 | 0.020 | – | – | – | – | – | – | – | – |
| | Accept | 1 | 1 | 0 | 1 | 0 | 0 | 0 | – | – | – | – | – | – | – | – |
| LSTM | $\Delta\mathcal{J}^t$ | 0.120 | 0.015 | 0.034 | 0.022 | 0.045 | -0.023 | -0.029 | -0.021 | – | – | – | – | – | – | – |
| | $\Delta\Gamma^{(t)}$ | 0.066 | 0.030 | 0.017 | 0.020 | 0.020 | 0.006 | 0.003 | 0.001 | – | – | – | – | – | – | – |
| | Accept | 1 | 1 | 1 | 1 | 1 | 0 | 0 | 0 | – | – | – | – | – | – | – |
| Transformer | $\Delta\mathcal{J}^t$ | 0.161 | 0.119 | 0.007 | 0.020 | 0.008 | 0.035 | 0.004 | 0.006 | -0.004 | 0.001 | 0.010 | 0.015 | -0.004 | -0.003 | -0.018 |
| | $\Delta\Gamma^{(t)}$ | 0.063 | 0.087 | 0.014 | 0.016 | 0.012 | 0.025 | 0.008 | 0.004 | -0.004 | 0.004 | 0.009 | 0.001 | 0.001 | -0.002 | -0.007 |
| | Accept | 1 | 1 | 1 | 1 | 1 | 1 | 1 | 1 | 0 | 1 | 1 | 1 | 0 | 0 | 0 |

TABLE IV
COMPREHENSIVE EXPERIMENTAL RESULTS OF IGADA-IOT ON 8 COMMONLY USED DOWNSTREAM MODELS.

| Model | Initialization | | | | Generated Samples | | | | Final Performance | | | | Cost | |
|---|---|---|---|---|---|---|---|---|---|---|---|---|---|---|
| | $\Gamma$ | $B^{(0)}$ | Test Acc | Macro-F1 | 0 | 1 | 2 | Final Size | Test Acc | Macro-F1 | $\Delta$Acc | $\Delta$F1 | Round | Time (s) |
| LR | 0.2564 | 159 | .745 ± .000 | .743 ± .000 | 66 | 18 | 75 | 159 | .773 ± .018 | .769 ± .031 | 0.0282 | 0.0260 | 3 | 0.2736 |
| RF | 0.2786 | 168 | .755 ± .055 | .737 ± .050 | 151 | 33 | 176 | 360 | .782 ± .055 | .765 ± .048 | 0.0273 | 0.0287 | 10 | 2.7111 |
| SVC | 0.2483 | 155 | .818 ± .009 | .817 ± .011 | 145 | 30 | 165 | 340 | .827 ± .018 | .829 ± .016 | 0.0091 | 0.0129 | 9 | 2.1624 |
| ANN | 0.2447 | 154 | .736 ± .064 | .740 ± .060 | 149 | 33 | 169 | 351 | .791 ± .064 | .790 ± .064 | 0.0545 | 0.0494 | 10 | 5.3641 |
| 1D-CNN | 0.2236 | 145 | .827 ± .064 | .828 ± .064 | 125 | 31 | 141 | 297 | .900 ± .027 | .898 ± .029 | 0.0727 | 0.0698 | 6 | 6.6182 |
| GRU | 0.2667 | 163 | .700 ± .055 | .697 ± .057 | 119 | 29 | 134 | 282 | .818 ± .027 | .819 ± .030 | 0.1182 | 0.1223 | 6 | 22.7636 |
| LSTM | 0.2650 | 162 | .718 ± .073 | .724 ± .073 | 158 | 36 | 177 | 371 | .818 ± .045 | .814 ± .042 | 0.1000 | 0.0900 | 7 | 23.5459 |
| Transformer | 0.2816 | 170 | .727 ± .245 | .730 ± .275 | 208 | 33 | 233 | 474 | .900 ± .036 | .901 ± .038 | 0.1727 | 0.1709 | 14 | 41.2351 |

Fig. 4 and Table III showed that $\Gamma^{(t)}$ continuously decreased over multiple rounds and eventually tended to stabilize. The gradual increase and final stabilization of the Test Acc indicated that $\Gamma^{(t)}$ was progressively reduced. Therefore, the HMGCS enhanced the targetedness and rationality of generated sample allocation, and the differentiated scheduling demands were satisfied. The rejected rounds were characterized by negative $\Delta\mathcal{J}^{(t)}$ or $\Delta\Gamma^{(t)}$. The marginal gain of the downstream models was weakened in these rounds. Therefore, the accuracy of augmentation decisions was enhanced, and the risks of under-augmentation and over-augmentation were mitigated. Fig. 5 further showed in detail the dynamic changes of the generator ratios over multiple rounds. It indicated that the HMGCS continuously adjusted the coordination between exploration and refinement according to $\Gamma^{(t)}$ and $\Delta\mathcal{J}^{(t)}$. The difference between the $B^{(0)}$ and "Final Size" indicated that generated samples were continuously accumulated on the basis of the launch injection. Specifically, the ratio of the Transformer's final $\Gamma^{(t)}$ to its $\Gamma$ was close to 5%, which provided empirical support for the chosen information-gap threshold. Therefore, capabilities of different generators were jointly utilized, and the targeted and quantitative compensation for information gaps was enhanced.

The quantitative results in Table IV showed that the IGADA-IoT improved Test Acc of the downstream models by an average of 7.28%, and increased the Macro-F1 by an average of 0.0713. The gains were more significant for complex temporal models such as the GRU, the LSTM, and the Transformer. The experimental results demonstrated the model generalizability and information compensation of the IGADA-IoT. In terms of computational cost, the IGADA-IoT required no more than 1 minute, indicating a favorable balance between performance and efficiency. Moreover, Fig. 6 showed the computational efficiency of the 8 downstream models. On the augmented dataset, the time costs on the 4 machine learning models were all below 1 s, and those on the 4 deep learning models were all below 5 s. The memory usage was below 2 MB for all models. It was worth noting that training the ImagenTime backbone on an RTX 3090 took 3 h 20 min, whereas sample generation took less than 0.1 min per round. Its end-to-end augmentation time was close to its backbone training time, and its memory usage on GPU was 439 MB. Since the generator ran offline, its relatively high computational cost did not affect downstream model deployment, computation, or inference. Therefore, the IGADA-IoT improved downstream model performance without introducing significant overhead to model training and deployment. Its low computational cost and high computational efficiency were beneficial for deployment and application in WSNs.

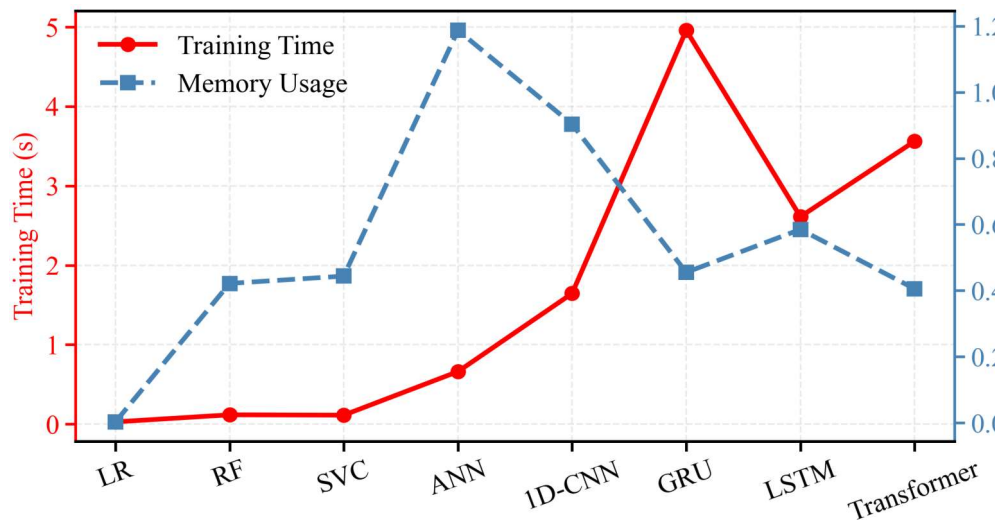

Fig. 6. Computational efficiency of 8 downstream models, including training time and memory usage.

To further assess the statistical reliability of the IGADA-IoT effects, we analyzed the 95% confidence intervals for Test Acc and Macro-F1 before and after data augmentation across ten runs with random seeds 42–51. Performance differences were evaluated using two-sided exact paired permutation tests with Holm correction for multiple comparisons. The two best-performing downstream models in Table IV, the 1D-CNN and the Transformer, were used for comparison to represent the downstream models. The 1D-CNN was used to represent a model with a relatively balanced trade-off between performance and efficiency, whereas the Transformer was used to represent the model with the best performance and the longest closed-loop augmentation rounds. The experimental results of the 1D-CNN and the Transformer were shown in Fig. 7. The experimental results in Fig. 7 showed that the confidence intervals for both the 1D-CNN and the Transformer narrowed after augmentation, and all adjusted p-values were below 0.05, indicating that the observed improvements in model performance were statistically significant.

*3) Comparisons with Existing Methods:* To validate the advancement of the IGADA-IoT in data augmentation, it was compared with existing data augmentation methods. The methods included ImagenTime [47] and other mainstream advanced generative models, including DS-Diffusion [44], Diffusion-TS [45], DiffTime [53], MG-TSD [46], KoVAE [42], and TimeGAN [54]. Advanced generation strategies were also used for comparison, including RIM [34], DTW [35], TSW [36], Non-SIM [11], SMOTE, and Borderline-SMOTE [37], [38]. Moreover, automated augmentation search and multi-generator methods were used for comparison, including AutoTCL [39], DoppelGANger [49], and ADAPTIVE [50]. Similarly, the 1D-CNN and the Transformer were used as representative downstream models. The generated sample sizes were equal to "Final Size" in Table IV for each downstream model. The comparisons were shown in Table V.

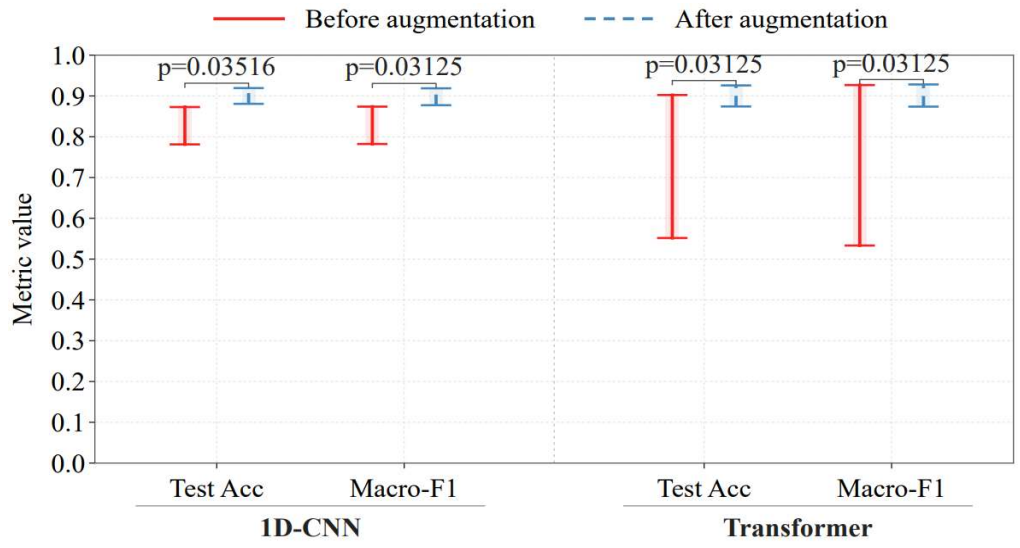

Fig. 7. Confidence intervals and significance tests for data augmentation.

TABLE V
COMPARISONS WITH EXISTING METHODS USING TWO REPRESENTATIVE DOWNSTREAM MODELS: 1D-CNN AND TRANSFORMER (BOLD INDICATES BEST PERFORMANCE).

| Method | 1D-CNN | | Transformer | |
|---|---|---|---|---|
| | Test Acc | Macro-F1 | Test Acc | Macro-F1 |
| **IGADA-IoT** | **.900 ± .027** | **.898 ± .029** | **.900 ± .036** | **.901 ± .038** |
| ImagenTime ($g_1$) | .845 ± .027 | .847 ± .032 | .809 ± .055 | .803 ± .059 |
| DS-Diffusion | .845 ± .009 | .844 ± .022 | .818 ± .036 | .801 ± .043 |
| Diffusion-TS | .836 ± .064 | .819 ± .050 | .800 ± .046 | .776 ± .047 |
| DiffTime | .836 ± .027 | .821 ± .031 | .791 ± .046 | .775 ± .056 |
| MG-TSD | .845 ± .027 | .834 ± .032 | .800 ± .036 | .783 ± .039 |
| KoVAE | .827 ± .055 | .813 ± .053 | .782 ± .046 | .775 ± .057 |
| TimeGAN | .809 ± .100 | .805 ± .104 | .736 ± .109 | .727 ± .088 |
| RIM ($g_2$) | .800 ± .055 | .799 ± .056 | .845 ± .055 | .843 ± .055 |
| TSW ($g_3$) | .836 ± .045 | .833 ± .052 | .800 ± .045 | .795 ± .052 |
| DTW ($g_4$) | .800 ± .073 | .803 ± .066 | .855 ± .036 | .851 ± .040 |
| Non-SIM | .818 ± .027 | .811 ± .041 | .791 ± .091 | .793 ± .039 |
| SMOTE | .782 ± .046 | .775 ± .057 | .727 ± .082 | .700 ± .082 |
| Borderline-SMOTE | .791 ± .063 | .786 ± .055 | .791 ± .064 | .782 ± .061 |
| AutoTCL | .845 ± .036 | .832 ± .037 | .809 ± .027 | .792 ± .032 |
| DoppelGANger | .809 ± .091 | .790 ± .093 | .800 ± .109 | .793 ± .113 |
| ADAPTIVE | .855 ± .091 | .842 ± .090 | .819 ± .091 | .801 ± .095 |

Experimental results in Table V showed that the IGADA-IoT achieved the best performance on both the 1D-CNN and the Transformer. Compared with the existing methods, the IGADA-IoT improved Test Acc and Macro-F1 by 8.90% and 9.81% on average on the two representative downstream models, respectively. The better performance of the downstream models indicated that the IGADA-IoT provided more stable and greater model performance gains. Therefore, the automatic data augmentation of the IGADA-IoT was demonstrated to compensate for the information gap more effectively, thereby improving the performance and generalizability of downstream models. The advancement of the IGADA-IoT in data augmentation was demonstrated.

*4) Comparisons on Public IoT Sensor Datasets:* To validate the generalizability of the IGADA-IoT, we also conducted experiments on public IoT sensor datasets. Five

TABLE VI

COMPARISONS ON REPRESENTATIVE IOT SENSOR DATASETS IN UCR ARCHIVE (BOLD INDICATES BEST PERFORMANCE).

| Model | Dataset | Class | $\Gamma$ | Final Size | Test Acc | Augmented Test Acc | $\Delta$Acc | ImagenTime ($g_1$) | RIM ($g_2$) | TSW ($g_3$) | DTW ($g_4$) |
|---|---|---|---|---|---|---|---|---|---|---|---|
| 1D-CNN | Earthquakes | 2 | 0.439 | 238 | .255±.009 | **.468±.191** | 0.213 | .388±.137 | .424±.173 | .432±.180 | .446±.194 |
| | FreezerRegularTrain | 2 | 0.262 | 139 | .761±.025 | **.969±.100** | 0.208 | .780±.141 | .760±.006 | .766±.016 | .760±.023 |
| | InsectWingbeatSound | 11 | 0.321 | 342 | .390±.017 | **.425±.016** | 0.035 | .403±.030 | .405±.028 | .419±.024 | .419±.016 |
| | ItalyPowerDemand | 2 | 0.108 | 104 | .933±.023 | **.950±.035** | 0.017 | .943±.029 | .916±.020 | .939±.024 | .937±.020 |
| | Trace | 4 | 0.214 | 154 | .810±.040 | **1.000±.000** | 0.190 | .960±.035 | .810±.169 | .900±.080 | .930±.047 |
| Transformer | Earthquakes | 2 | 0.401 | 221 | .252±.237 | **.662±.079** | 0.410 | .604±.029 | .649±.017 | .590±.038 | .583±.057 |
| | FreezerRegularTrain | 2 | 0.077 | 169 | .923±.068 | **.991±.018** | 0.068 | .979±.028 | .990±.026 | .924±.036 | .986±.015 |
| | InsectWingbeatSound | 11 | 0.247 | 342 | .558±.028 | **.576±.016** | 0.018 | .573±.017 | .560±.024 | .571±.018 | .565±.025 |
| | ItalyPowerDemand | 2 | 0.047 | 102 | .936±.041 | **.960±.044** | 0.024 | .940±.022 | .954±.028 | .960±.020 | .720±.195 |
| | Trace | 4 | 0.268 | 154 | .810±.024 | **.899±.018** | 0.089 | .812±.028 | .824±.016 | .820±.019 | .890±.010 |

representative IoT sensor datasets from the UCR Archive, namely "Earthquakes", "FreezerRegularTrain", "InsectWingbeatSound", "ItalyPowerDemand", and "Trace", were used for comparisons [55]. The datasets covered different sensor modalities, class scales, and temporal complexities, and thus more comprehensively represented classification scenarios of multiple sensors in WSNs. The individual generators in the IGADA-IoT represented a state-of-the-art generative model and advanced generation strategies. Therefore, the comparisons between the IGADA-IoT and its individual generators, which served as representative baselines, were used to validate the generalizability. Similarly, only $\mathcal{D}_{tr}$ were used for data augmentation. 1D-CNN and Transformer were used for comparisons. Grid search and K-fold cross-validation were also used. The main metrics were shown in Table VI.

The experimental results showed that the IGADA-IoT achieved the best or tied-best performance on all 5 public datasets, with average accuracy gains of 13.26% and 12.18% for the 1D-CNNs and the Transformers, respectively. The overall average improvement was 12.72%. Moreover, the data augmentation performance of the IGADA-IoT outperformed that of the individual generators. The experiments on public datasets showed that the IGADA-IoT could stably improve the classification performance in environments involving multiple IoT sensors. The generalizability of the IGADA-IoT was further demonstrated.

*5) Effects of Final Size on Downstream Performance:* To evaluate the effectiveness of the IGADA-IoT in mitigating under-augmentation and over-augmentation, we compared 5 different Final Sizes in data augmentation: $0.5M$, $0.75M$, $M$, $1.5M$, and $2M$, where $M$ denoted the Final Size for each model accepted by the IGADA-IoT. Specifically, $0.5M$ and $0.75M$ were used to simulate under-augmentation, whereas $1.5M$ and $2M$ were used to simulate over-augmentation. The experimental results were shown in Table VII. Table VII showed that all the downstream models achieved their highest Test Acc and Macro-F1 at $M$, whereas both smaller and larger sizes yielded lower mean performance. Therefore, the effectiveness of the IGADA-IoT in mitigating both issues was empirically supported within the evaluated range.

Furthermore, we analyzed the sensitivity of $\gamma_c^{(t)}$ to its component weights using 1D-CNN and Transformer. In the configurations (W01–W07), the weights of $h_{c,\text{size}}^{(t)}$ and $h_{c,\text{dist}}^{(t)}$ were kept equal and increased from 0.10 to 0.40 in increments of 0.05, while the weights of $h_{c,\text{bdry}}^{(t)}$ and $h_{c,\text{unc}}^{(t)}$ were kept equal and decreased from 0.40 to 0.10, ensuring that the four weights summed to one. W04 assigned a weight of 0.25 to each component. Mean values of Test Acc and Macro-F1 were used for evaluation. As shown in Fig. 8, the equal-weight configuration achieved the best performance among the tested settings, supporting the choice of equal weights.

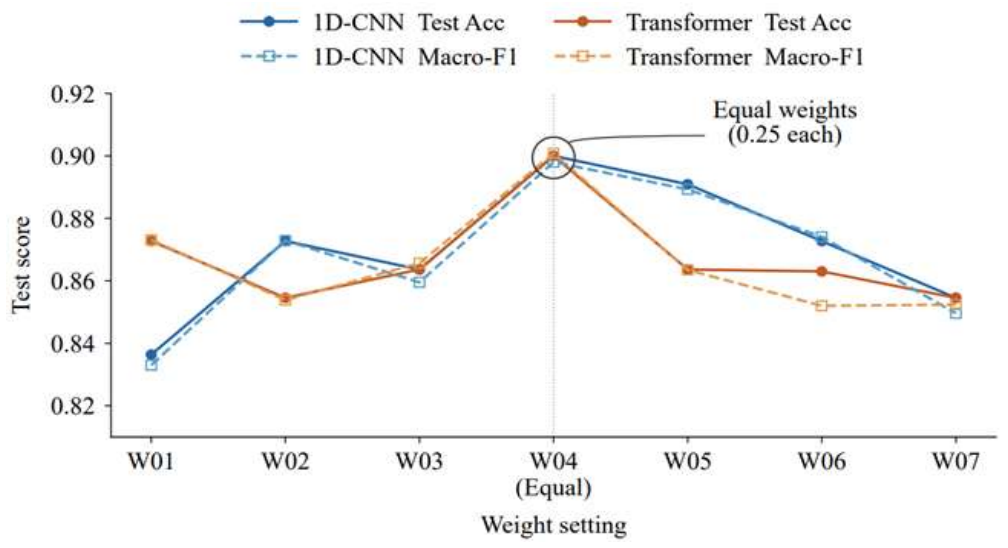

Fig. 8. Weight sensitivity analysis of $\gamma_c^{(t)}$.

## D. Ablation Study

To verify the contribution of each component in the IGADA-IoT, this subsection conducted a detailed ablation study. Similarly, the 1D-CNN and the Transformer were used. First, the hierarchical multi-generator architecture was ablated. The 4 single individual generators were used to replace the hierarchical multi-generator architecture in the IGADA-IoT with the same "Final Size." Whether the performance gain came from the complementarity among multiple generators was verified. Second, the GCM was ablated. $\boldsymbol{A}$ was replaced by tensors with constant values of 0.3, 0.5, and 1, respectively. The homogeneous tensors represented homogeneous multiple generators. The role of differentiated generator capability in collaboration and scheduling was verified. Third, the class-level decision in the HMGCS was ablated. $P_{\text{cls}}^{(t)}(c)$ was replaced with a uniform distribution to verify the guidance of the HMGCS on class allocation over multiple rounds. Fourth, the generator-level decision in the HMGCS was ablated. $P_{\text{gen}}^{(t)}(g \mid c)$ was replaced with a uniform distribution to verify the dependence of $E_c^{(t)}$ and $R_c^{(t)}$ on the capabilities of each generator over multiple rounds. Fifth, the budget-level decision in the HMGCS was ablated. $n_{g,c}^{(t)}$ was removed, and the dynamic update of $B^{(t+1)}$ was further removed. The guidance of the HMGCS on budget allocation across generators over multiple rounds was verified. Sixth, the IGMP-EC was ablated. The IGMP-E and IGMP-C were removed separately, and then the IGMP-EC was removed entirely. When the IGMP-E was removed, the guidance from the information gap and model performance disappeared. When the IGMP-C was removed, the closed loop of launch injection, refinement adjustment, and multi-round accept-or-reject was no longer performed. They were also removed simultaneously, so that the automatic data augmentation degraded into an open-loop process with a fixed quantity and full acceptance. The role of the IGMP-EC in suppressing ineffective augmentation and avoiding over-augmentation was verified. Finally, the guidance of the information gap and the constraint of model performance in the IGMP-EC were further verified. The metric set that represented model performance was denoted as $\mathcal{MP}$, including $\mathrm{MacroF1}^{(t)}$, $\mathrm{Acc}^{(t)}$, $\mathrm{MacroPre}^{(t)}$, and $\mathrm{MacroRec}^{(t)}$. $\Gamma^{(t)}$ and $\mathcal{MP}$ were ablated.

TABLE VII
PERFORMANCE UNDER DIFFERENT FINAL SIZES (BOLD INDICATES BEST PERFORMANCE).

| Model | Metric | Final Size | | | | |
|---|---|---|---|---|---|---|
| | | $0.5M$ | $0.75M$ | $M$ | $1.5M$ | $2M$ |
| LR | Test Acc | $.746 \pm .021$ | $.747 \pm .019$ | $\mathbf{.773 \pm .018}$ | $.745 \pm .018$ | $.752 \pm .018$ |
| | Macro-F1 | $.746 \pm .022$ | $.747 \pm .020$ | $\mathbf{.769 \pm .031}$ | $.744 \pm .019$ | $.751 \pm .017$ |
| RF | Test Acc | $.731 \pm .022$ | $.782 \pm .064$ | $\mathbf{.782 \pm .055}$ | $.735 \pm .027$ | $.745 \pm .018$ |
| | Macro-F1 | $.712 \pm .017$ | $.765 \pm .067$ | $\mathbf{.765 \pm .048}$ | $.720 \pm .025$ | $.727 \pm .021$ |
| SVC | Test Acc | $.787 \pm .016$ | $.792 \pm .023$ | $\mathbf{.827 \pm .018}$ | $.802 \pm .020$ | $.797 \pm .023$ |
| | Macro-F1 | $.787 \pm .016$ | $.791 \pm .023$ | $\mathbf{.829 \pm .016}$ | $.801 \pm .020$ | $.797 \pm .022$ |
| ANN | Test Acc | $.769 \pm .046$ | $.758 \pm .050$ | $\mathbf{.791 \pm .064}$ | $\mathbf{.791 \pm .064}$ | $.776 \pm .050$ |
| | Macro-F1 | $.768 \pm .047$ | $.758 \pm .053$ | $\mathbf{.790 \pm .064}$ | $\mathbf{.790 \pm .064}$ | $.773 \pm .052$ |
| 1D-CNN | Test Acc | $.815 \pm .020$ | $.829 \pm .019$ | $\mathbf{.900 \pm .027}$ | $.849 \pm .016$ | $.835 \pm .021$ |
| | Macro-F1 | $.816 \pm .021$ | $.828 \pm .018$ | $\mathbf{.898 \pm .029}$ | $.847 \pm .015$ | $.831 \pm .022$ |
| GRU | Test Acc | $.759 \pm .023$ | $.778 \pm .022$ | $\mathbf{.818 \pm .027}$ | $.791 \pm .017$ | $.803 \pm .017$ |
| | Macro-F1 | $.757 \pm .022$ | $.776 \pm .021$ | $\mathbf{.819 \pm .030}$ | $.786 \pm .018$ | $.797 \pm .017$ |
| LSTM | Test Acc | $.784 \pm .025$ | $.791 \pm .018$ | $\mathbf{.818 \pm .045}$ | $.814 \pm .010$ | $.805 \pm .017$ |
| | Macro-F1 | $.783 \pm .025$ | $.789 \pm .020$ | $\mathbf{.814 \pm .042}$ | $.810 \pm .009$ | $.801 \pm .017$ |
| Transformer | Test Acc | $.835 \pm .021$ | $.834 \pm .020$ | $\mathbf{.900 \pm .036}$ | $.839 \pm .016$ | $.835 \pm .024$ |
| | Macro-F1 | $.833 \pm .021$ | $.831 \pm .020$ | $\mathbf{.901 \pm .038}$ | $.835 \pm .016$ | $.831 \pm .024$ |

Table VIII demonstrated the contribution of each component in detail. First, the ablation of the hierarchical multi-generator architecture indicated that the hierarchical multi-generator architecture outperformed any individual generator with the same "Final Size." Compared with the individual generators of the IGADA-IoT, the average Test Acc was improved by 7.63%. The performance gain originated from the complementarity among multiple generators in feature exploration and boundary refinement. Second, the model performance declined to different degrees after $\boldsymbol{A}$ was replaced with homogeneous tensors. It demonstrated that tensorizing the differentiated capability of generators was the foundation of the HMGCS. Third, the model performance decreased after $P_{\text{cls}}^{(t)}(c)$ was replaced with a uniform distribution. It demonstrated that the class-level decision led by the HMGCS achieved the class-level scheduling distribution precisely. Fourth, the performance degradation after $P_{\text{gen}}^{(t)}(g \mid c)$ was replaced with a uniform distribution demonstrated the dependence of $E_c^{(t)}$ and $R_c^{(t)}$ on the capabilities of each generator. The dependence was realized through the generator-level decision. Fifth, removing $n_{g,c}^{(t)}$ caused budget allocation of generation to become rigid. It demonstrated that the budget-level decision was critical for maximizing the marginal gain of generated data. The ablation of the class-level decision, the generator-level decision, and the budget-level decision in the HMGCS also demonstrated that the HMGCS enhanced the targetedness and rationality of generated sample allocation, and differentiated scheduling demands were satisfied. Sixth, the absence of the IGMP-EC caused the IGADA-IoT to degrade into an open-loop mode, which led to a clear performance decline. It demonstrated the irreplaceable role of the IGMP-EC in multi-round augmentation decisions, which mitigated the risks of under-augmentation and over-augmentation. Finally, the model performance declined after $\Gamma^{(t)}$ or $\mathcal{MP}$ in $\mathcal{J}^{(t)}$ was removed. It demonstrated that the guidance of the information gap and the constraint of model performance were both indispensable in the IGMP-EC. The degradation caused by removing $\Gamma^{(t)}$ was more obvious, which further demonstrated the dominant guiding role of $\Gamma^{(t)}$ in the IGADA-IoT. Therefore, the ablation study systematically demonstrated the contribution of each component in the IGADA-IoT. It also demonstrated that the advantage of the IGADA-IoT in data augmentation did not originate from the superposition of an individual component, but from the deep coupling of the hierarchical multi-generator architecture, the GCM, the HMGCS, and the IGMP-EC.

Furthermore, to validate the benefits of multi-generator collaboration and adaptive scheduling under a common closed loop, we evaluated each generator individually within the same closed loop and included a fixed mixture of the four generators without adaptive scheduling. All variants followed the same validation, retry, acceptance, and stopping rules, as shown in Table IX. Notably, these comparisons differed from those in Table V, where the final number of generated samples was kept identical across methods. Each generator shared the same evaluation, acceptance, and stopping rules. Table IX showed that the IGADA-IoT achieved the highest Test Acc and Macro-F1 among the compared variants. The results demonstrated that the HMGCS provided additional performance gains by adaptively allocating samples across classes and generators within the closed loop. Their integration combined adaptive scheduling with validation-based candidate evaluation, acceptance or rejection, and stopping, supporting effective multi-round data augmentation.

TABLE VIII
ABLATION STUDY (BOLD INDICATES BEST PERFORMANCE).

| Ablation Target | | 1D-CNN | | Transformer | |
|---|---|---|---|---|---|
| | | Test Acc | Macro-F1 | Test Acc | Macro-F1 |
| **No Ablation** | | **.900 ± .027** | **.898 ± .029** | **.900 ± .036** | **.901 ± .038** |
| Architecture | $g_1$ | .845 ± .027 | .847 ± .032 | .809 ± .055 | .803 ± .059 |
| | $g_2$ | .800 ± .055 | .799 ± .056 | .845 ± .055 | .843 ± .055 |
| | $g_3$ | .836 ± .045 | .833 ± .052 | .800 ± .045 | .795 ± .052 |
| | $g_4$ | .800 ± .073 | .803 ± .066 | .855 ± .036 | .851 ± .040 |
| GCM | $\boldsymbol{A} \leftarrow \mathbf{0.3}$ | .873 ± .045 | .873 ± .050 | .891 ± .064 | .890 ± .064 |
| | $\boldsymbol{A} \leftarrow \mathbf{0.5}$ | .827 ± .064 | .826 ± .065 | .882 ± .055 | .881 ± .054 |
| | $\boldsymbol{A} \leftarrow \mathbf{1.0}$ | .791 ± .073 | .789 ± .075 | .827 ± .045 | .826 ± .047 |
| HMGCS | Class-level: $P^{(t)}_{\text{cls}}(c)$ | .873 ± .027 | .870 ± .027 | .855 ± .036 | .855 ± .035 |
| | Generator-level: $P^{(t)}_{\text{gen}}(g \mid c)$ | .855 ± .018 | .855 ± .023 | .873 ± .055 | .873 ± .060 |
| | Budget-level: $n^{(t)}_{g,c}$ | .864 ± .027 | .864 ± .034 | .882 ± .055 | .880 ± .054 |
| IGMP-EC | IGMP-E | .882 ± .027 | .878 ± .053 | .855 ± .027 | .844 ± .033 |
| | IGMP-C | .864 ± .018 | .856 ± .035 | .855 ± .018 | .847 ± .030 |
| | IGMP-EC | .764 ± .118 | .751 ± .123 | .773 ± .109 | .757 ± .101 |
| $\mathcal{J}^{(t)}$ | $\Gamma^{(t)}$ | .846 ± .018 | .834 ± .035 | .827 ± .018 | .819 ± .017 |
| | $\mathcal{MP}$ | .836 ± .009 | .822 ± .020 | .836 ± .009 | .829 ± .019 |

TABLE IX
COMPARISON UNDER A COMMON CLOSED LOOP (BOLD INDICATES BEST PERFORMANCE).

| Method | 1D-CNN | | Transformer | |
|---|---|---|---|---|
| | Test Acc | Macro-F1 | Test Acc | Macro-F1 |
| IGADA-IoT | **.900 ± .027** | **.898 ± .029** | **.900 ± .036** | **.901 ± .038** |
| No augmentation | .827 ± .064 | .828 ± .064 | .727 ± .245 | .730 ± .275 |
| ImagenTime ($g_1$) | .864 ± .009 | .852 ± .018 | .845 ± .036 | .832 ± .037 |
| RIM ($g_2$) | .846 ± .014 | .844 ± .015 | .836 ± .018 | .836 ± .017 |
| TSW ($g_3$) | .861 ± .015 | .859 ± .016 | .841 ± .016 | .839 ± .017 |
| DTW ($g_4$) | .855 ± .022 | .854 ± .023 | .840 ± .021 | .838 ± .021 |
| Fixed mix | .858 ± .024 | .856 ± .024 | .829 ± .023 | .829 ± .022 |

## *E. Real-world Deployments*

To further verify the accuracy and deployment feasibility of the IGADA-IoT, this subsection showed its real-world deployments in WSNs. “Fuxi” was deployed on local farms to continuously acquire data, as shown in Fig. 9. The farms were located in northeastern China. “Fuxi” operated at a normal voltage of 12 V and a current of 1.12 A. The energy consumption of a single sampling event was 3.73 mWh. The energy savings were based primarily on sensing operations and did not account for communication or inference costs [56]. Considering the balance between model performance and computational efficiency, the 1D-CNN was used as the representative downstream model and was deployed on “Fuxi”. In the real-world deployments, the 1D-CNN trained without the IGADA-IoT augmentation (denoted as $\mathscr{G}_{\text{N}}$) and the 1D-CNN trained with augmentation (denoted as $\mathscr{G}_{\text{A}}$) were used for sampling frequency decision-making for energy optimization. 6 days were used for energy optimization validation, including 2025.06.16, 2025.08.20, 2025.08.29, 2025.09.02, 2025.09.18, and 2025.09.19. The dates covered multiple non-consecutive days with different environments, which reduced the influence of incidental fluctuations on the validation. The results of real-world deployments were shown in Table X.

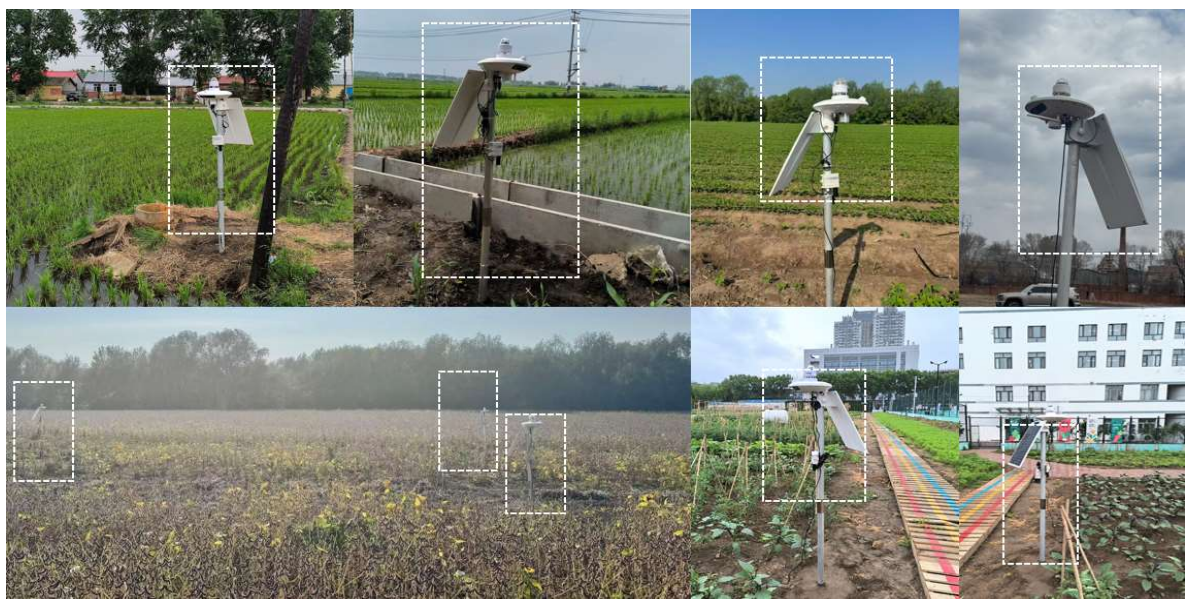

Fig. 9. Real-world deployments: “Fuxi” on local farms.

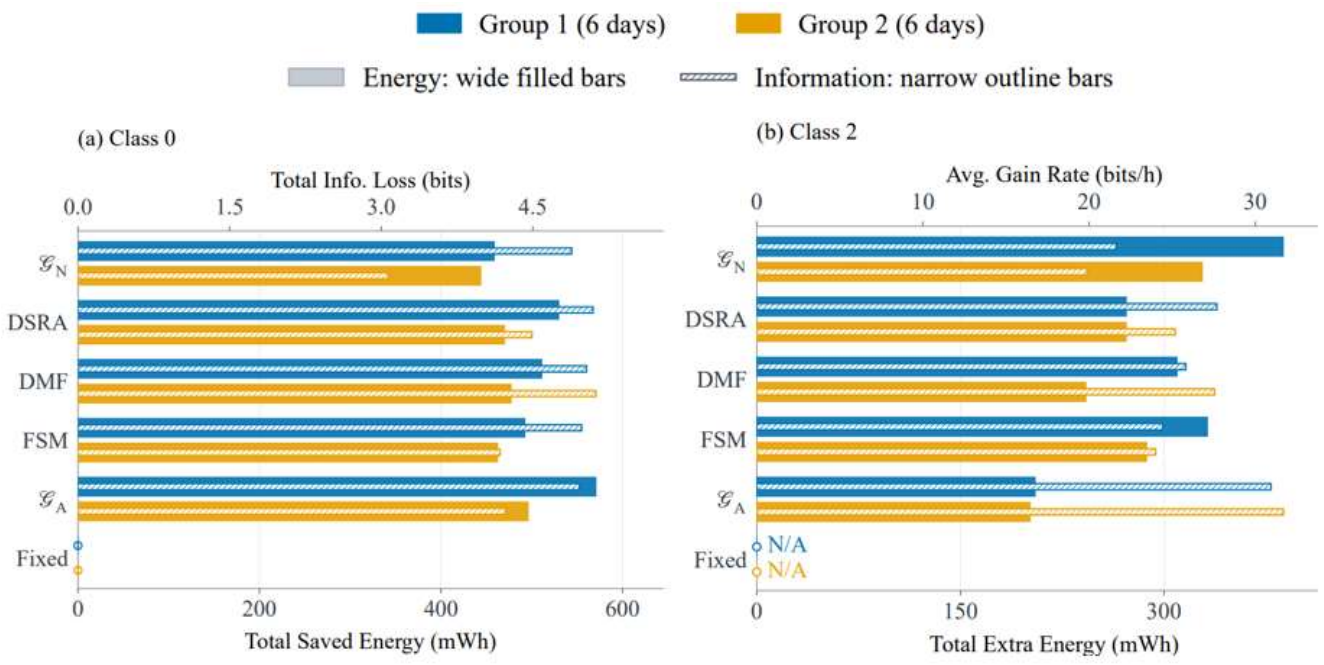

Fig. 10. Comparisons across two additional groups.

In Table X, when the class was 0, "Class=0 Dur." denoted the total duration of environmentally stable periods, "Saved Samplings" denoted the reduced number of samplings due to lowering the sampling frequency, "Info. Loss" denoted the information loss caused by lowering the sampling frequency, and "Saved Energy" denoted the corresponding saved energy consumption. When the class was 2, "Class=2 Dur." denoted the total duration of periods with drastic environmental changes, "Extra Samplings" denoted the increased number of samplings due to enriching the information content, "Extra Energy" denoted the corresponding extra energy consumption, "Extra Info." denoted the additional information obtained by increasing the sampling frequency, and "Avg. Gain Rate" denoted the information gain efficiency per unit time. "Exp. Dur." denoted the total duration of the experiment. "Info. Loss" was computed from standardized reconstruction residuals at omitted timestamps using linear interpolation of the sparse sequence. A base-2 log-determinant measure was applied to

TABLE X
REAL-WORLD DEPLOYMENTS OF NO AUGMENTATION AND IGADA-IOT AUGMENTATION IN STABLE ($Class = 0$) AND UNSTABLE ($Class = 2$) ENVIRONMENTS.

**Panel A. Energy optimization in $Class = 0$ (stable environments).**

| Group | Date | $Class = 0$ Dur. (h) | Saved Samplings | Info. Loss (bits) | Saved Energy (mWh) | Exp. Dur. (h) |
|---|---|---|---|---|---|---|
| $\mathscr{G}_{\mathrm{N}}$ | 2025.06.16 | – | – | – | – | 12.0650 |
| | 2025.08.20 | 0.8750 | 7 | 0.0162 | 26.11 | 16.1239 |
| | 2025.08.29 | 0.4769 | 4 | 0.6377 | 14.92 | 15.8033 |
| | 2025.09.02 | 7.1210 | 54 | 0.5450 | 201.42 | 23.9917 |
| | 2025.09.18 | 5.0933 | 38 | 0.0929 | 141.74 | 23.9828 |
| | 2025.09.19 | 3.4621 | 26 | 2.8221 | 96.98 | 7.9531 |
| $\mathscr{G}_{\mathrm{A}}$ | 2025.06.16 | – | – | – | – | 12.0650 |
| | 2025.08.20 | 0.8967 | 7 | 0.0419 | 26.11 | 16.1239 |
| | 2025.08.29 | 0.4703 | 4 | 0.8833 | 14.92 | 15.8033 |
| | 2025.09.02 | 8.6081 | 65 | 0.8108 | 242.45 | 23.9917 |
| | 2025.09.18 | 5.5336 | 42 | 1.0016 | 156.66 | 23.9828 |
| | 2025.09.19 | 4.6306 | 35 | 2.7101 | 130.55 | 7.9531 |

*Summary:* Energy per sampling = 3.73 mWh. No Augmentation: total Saved Samplings = 129, total Saved Energy = 481.17 mWh. IGADA-IoT Augmentation: total Saved Samplings = 153, total Saved Energy = 570.69 mWh.

**Panel B. Information gains in $Class = 2$ (unstable environments).**

| Group | Date | $Class = 2$ Dur. (h) | Extra Samplings | Extra Energy (mWh) | Extra Info. (bits) | Avg. Gain Rate (bits/h) | Exp. Dur. (h) |
|---|---|---|---|---|---|---|---|
| $\mathscr{G}_{\mathrm{N}}$ | 2025.06.16 | 0.0510 | 1 | 3.73 | 0.5532 | 10.8471 | 12.0650 |
| | 2025.08.20 | 0.0286 | 1 | 3.73 | 0.1625 | 5.6818 | 16.1239 |
| | 2025.08.29 | 1.1089 | 17 | 63.41 | 17.9160 | 16.1566 | 15.8033 |
| | 2025.09.02 | 3.9942 | 60 | 223.80 | 40.1198 | 10.0445 | 23.9917 |
| | 2025.09.18 | 1.0270 | 15 | 55.95 | 62.3635 | 60.7240 | 23.9828 |
| | 2025.09.19 | – | – | – | – | – | 7.9531 |
| $\mathscr{G}_{\mathrm{A}}$ | 2025.06.16 | 0.4753 | 7 | 26.11 | 5.4154 | 11.3942 | 12.0650 |
| | 2025.08.20 | 0.3886 | 5 | 18.65 | 3.0190 | 7.7687 | 16.1239 |
| | 2025.08.29 | 0.5383 | 8 | 29.84 | 15.0556 | 27.9670 | 15.8033 |
| | 2025.09.02 | 1.5800 | 23 | 85.79 | 35.3179 | 22.9912 | 23.9917 |
| | 2025.09.18 | 0.6636 | 10 | 37.30 | 55.3242 | 83.4469 | 23.9828 |
| | 2025.09.19 | – | – | – | – | – | 7.9531 |

*Summary:* No Augmentation: total Extra Samplings = 94, total Extra Energy = 350.62 mWh. total Extra Info. = 121.1150 bits, mean Avg. Gain Rate = 20.6908 bits/h. IGADA-IoT Augmentation: total Extra Samplings = 53, total Extra Energy = 197.69 mWh. total Extra Info. = 114.1321 bits, mean Avg. Gain Rate = 30.7136 bits/h.

their regularized covariance. For a single omitted observation, a channelwise squared-residual approximation was used. Results were reported in bits. "Extra Info." was computed by applying the same measure to the covariance of first differences of the reconstruction residuals. The result was converted to bits/h using the mean sampling interval. The resulting rate was multiplied by the interval duration and summed across valid class-2 intervals, with totals reported in bits. For reproducibility, calculations for both classes 0 and 2 required 5 full observations and 3 sparse observations, including context. Invalid intervals were excluded. The summary further showed a comparison of the variables for each group. The real-world deployments were evaluated in class 0 and class 2.

The experimental results in Table X showed that the IGADA-IoT increased the total "Saved Energy" by 18.6% while incurring only minimal information loss in stable environments, and improved mean "Avg. Gain Rate" by 48.44% while reducing the total "Extra Energy" by 43.62% in unstable environments. The total energy consumption was reduced by 7.62%. When the class was 0, $\mathscr{G}_{\mathrm{A}}$ achieved more "Saved Samplings", which further reduced the total energy consumption. $\mathscr{G}_{\mathrm{A}}$ could more stably identify stable periods and execute lower-frequency sampling. When the class was 2, although the total "Extra Info." decreased, the mean "Avg. Gain Rate" increased. The IGADA-IoT did not simply extend the total duration of high-frequency sampling. Instead, it more selectively concentrated high-frequency sampling on periods of abrupt change with higher informational value. In addition, "Class=0 Dur." in stable environments and "Class=2 Dur." in unstable environments indicated that the IGADA-IoT not only identified as many stable and unstable environments as possible, but also reduced the risk of misclassification. The energy optimization was thus more precise. Higher classification accuracy indicated fewer incorrect assignments of sampling intervals to environmental states. More precise sampling decisions helped reduce unnecessary acquisitions during stable periods and allocate high-frequency sampling to periods of rapid environmental change. The results of the real-world deployments demonstrated that the IGADA-IoT not only improved the performance of the downstream sampling-frequency decision models in offline experiments, but also provided better energy optimization performance in real-world WSNs. The energy consumption of IoT sensors was reduced while the information content of the acquired data was enriched. In the acquired datasets of WSNs with limited size and class imbalance, the IGADA-IoT could support the training of complex models to achieve precise energy optimization in dynamic environments.

Furthermore, comparisons were conducted in two additional six-day experimental groups. Besides $\mathscr{G}_{\mathrm{N}}$ and $\mathscr{G}_{\mathrm{A}}$, the comparison included the dynamic sampling rate algorithm (DSRA, representing adaptive sampling) [20], dynamic monitoring frequency (DMF, representing threshold-based sampling) [22], a finite state machine (FSM)-based energy-aware sensing method [21], and a fixed-frequency baseline with a 4-min sampling interval. For Class 0, we compared total Saved Energy and total Info. Loss. For Class 2, we compared total Extra Energy and mean Avg. Gain Rate. All comparisons used consistent energy accounting and identical information-metric definitions. The results were presented in Fig. 10. Across deployments conducted in different regions and at different times, $\mathscr{G}_{\mathrm{A}}$ achieved the lowest mean daily Extra Energy of $34.19\,\mathrm{mWh}$ (95% CI: $[1.22, 67.17]$) and $33.57\,\mathrm{mWh}$ (95% CI: $[3.66, 63.48]$), with the lowest corresponding sample variances among the compared methods ($987.35$ and $812.51\,\mathrm{mWh}^2$, respectively). $\mathscr{G}_{\mathrm{A}}$ achieved a more favorable balance between sampling energy consumption, information retention, and information gain than the other evaluated methods. More information was acquired from the environment with lower energy consumption. Therefore, the accuracy and deployment feasibility of the IGADA-IoT were further enhanced.

## V. CONCLUSION

In this study, the main objective is to develop an automatic data augmentation method for IoT sensor energy optimization in WSNs. Therefore, the IGADA-IoT with hierarchical multi-generator collaboration and scheduling over multiple rounds is proposed. Capabilities of different generators are jointly utilized to reduce the information gaps. The IGADA-IoT improves the average accuracy of multiple downstream models. Compared with advanced data augmentation methods and the individual generators, the average accuracy is also improved. Furthermore, public IoT sensor datasets from the UCR Archive and real-world deployments demonstrate the accuracy and deployment feasibility of the proposed method.

In future research, we will explore finer-grained multiple-objective optimization. Energy consumption, information acquisition efficiency, inference latency, and device resource usage will be jointly incorporated into the augmentation closed loop to meet differentiated requirements under different edge deployments. Each generator's validation utility will also be incorporated into its capability assessment to more directly reflect the value of its generated data for downstream tasks and improve multi-generator scheduling. In addition, "Fuxi" will also be extended to WSNs that include large-scale IoT sensor deployments across diverse regions and environmental settings. Multi-node collaborative sensing and cross-device automatic data augmentation will be explored to mine the complementary information among devices in the network. Furthermore, automatic data augmentation methods will investigate deeper integration with prior knowledge such as physical laws, domain knowledge, and temporal structures. The realism and logical consistency of generated data will be improved. The cross-domain transferability of the IGADA-IoT across multiple IoT domains will also be explored. A more general, intelligent, and resource-aware end-to-end IoT sensing system will provide broader support for its large-scale industrial application.

## Acknowledgments

This work was supported by the Key Research and Development Program of Heilongjiang Province (Grant No. 2023ZX01A24), the National Natural Science Foundation of China (Grant No. 62350710797 and No. 62103161), the Key Research and Development Program of Heilongjiang Province (Grant No. JD2023GJ01-01), and the Project of Laboratory of Advanced Agricultural Sciences, Heilongjiang Province (Grant No. ZY04JD05-010).

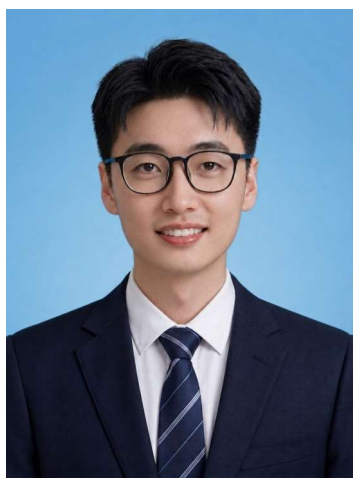

**Mingchun Sun** received the B.Eng. and M.Eng. degrees in control science and engineering from the University of Electronic Science and Technology of China, Chengdu, China, in 2018 and 2021, respectively.

He is currently pursuing the Ph.D. degree with Harbin Institute of Technology, Harbin, China. His current research interests include controllable data augmentation and deep learning on datasets with limited sample size.

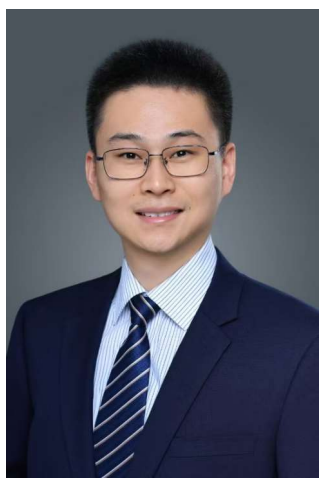

**Rongqiang zhao** received the Ph.D. degree in control science and engineering from the Harbin Institute of Technology, Harbin, China, in 2018.

He is currently an Associate Professor with the Faculty of Computing, Harbin Institute of Technology, Harbin 150001, China, and also with the State Key Laboratory of Smart Farm Technologies and Systems, Harbin Institute of Technology, Harbin 150001, China.

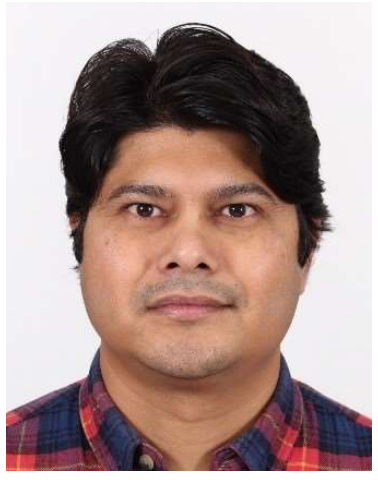

**Muhammad Abdul Munnaf** received a PhD in Bioscience Engineering from Ghent University, Belgium.

He is currently an Assistant Professor in Sensor Systems  Precision Agriculture at Agricultural Biosystems Engineering group in Wageningen University and Research, Netherlands. He is a member of IEEE (Circuits  Systems Society, Sensors  RFID councils) and the International Society of Precision Agriculture (ISPA).

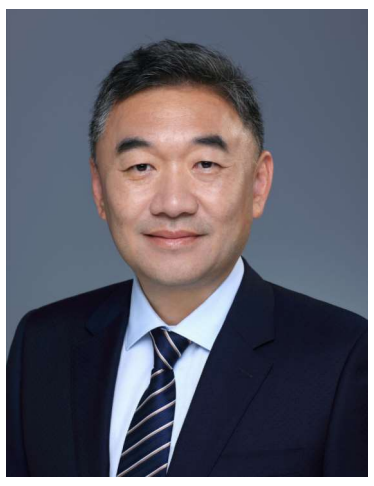

**Jie Liu** (Fellow, IEEE) received the Ph.D. degree in electrical engineering and computer science from the University of California at Berkeley, Berkeley, CA, USA, in 2001.

He is currently a Chair Professor with the Harbin Institute of Technology, Shenzhen, China. His research interests include artificial intelligence, control engineering, Internet of Things, and computer systems.